\documentclass{article}

\usepackage[final]{exait_2025}

\usepackage{algorithm}
\usepackage{algpseudocode}
\usepackage[utf8]{inputenc} 
\usepackage[T1]{fontenc}    
\usepackage{hyperref}       
\usepackage{url}            
\usepackage{booktabs}       
\usepackage{amsfonts}       
\usepackage{nicefrac}       
\usepackage{microtype}      
\usepackage{xcolor}         

\usepackage{graphicx} 
\usepackage{subcaption} 
\usepackage{amsmath}
\usepackage{amssymb}
\usepackage{textcomp}
\usepackage{gensymb}
\graphicspath{{images/}}
\usepackage{multirow}
\usepackage{arydshln}

\title{Neuro-Symbolic Financial Reasoning via Deterministic Fact Ledgers and Adversarial Low-Latency Hallucination Detector}

\author{%
  Pedram Agand\thanks{Research conducted in FactAI Lab available at \text{https://gofactai.com}} \\
    FactAI Lab,\\
  Vancouver, BC, Canada. \\
  \textit{pagand@gofactai.com}
}

\begin{document}

\maketitle

\begin{abstract}
Standard Retrieval-Augmented Generation (RAG) architectures fail in high-stakes financial domains due to two fundamental limitations: the inherent arithmetic incompetence of Large Language Models (LLMs) and the distributional semantic conflation of dense vector retrieval (e.g., mapping \textit{Net Income} to \textit{Net Sales} due to contextual proximity). In deterministic domains, a $99\%$ accuracy rate yields $0\%$ operational trust. To achieve zero-hallucination financial reasoning, we introduce the \textbf{Verifiable Numerical Reasoning Agent (VeNRA)}.
VeNRA shifts the RAG paradigm from retrieving probabilistic text to retrieving deterministic variables via a strictly typed Universal Fact Ledger (UFL). We mathematically bound this ledger using a novel \textbf{Double-Lock Grounding} algorithm. Coupled with deterministic Python execution, this neuro-symbolic routing compresses systemic hallucination rates to a near-zero $1.2\%$. Recognising that upstream parsing anomalies inevitably occur, we introduce the \textbf{VeNRA Sentinel}: a 3-billion parameter SLM trained to forensically audit candidate using a single-token inference budget with optional post-hoc reasoning. To train the Sentinel, we steer away from traditional hallucination datasets in favour of \textit{Adversarial Simulation}, programmatically sabotaging financial records to simulate \textit{Ecological Errors}. The compact Sentinel consequently outperforms $70$B$+$ frontier models in error detection. Through \textit{Loss Dilution} phenomenon in Reverse-CoT training, we present a novel \textbf{Micro-Chunking} loss algorithm to stabilise gradients under extreme verdict penalisation, yielding a $28{\times}$ latency speedup without sacrificing forensic rigor.
\end{abstract}

\section{Introduction}
\label{sec:intro}
\begin{figure*}[t]
\centering
\includegraphics[width=0.7\textwidth]{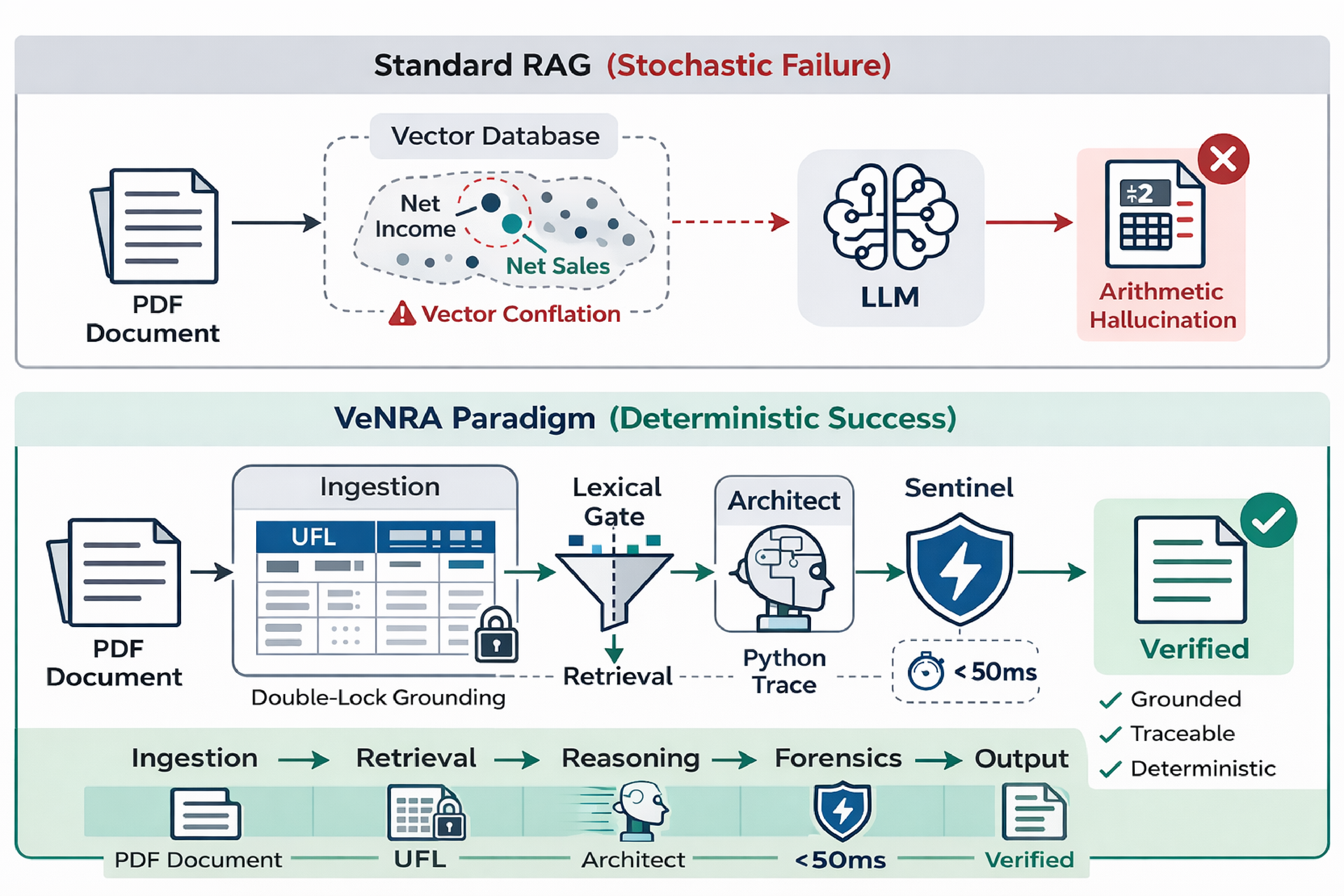}
\caption{\textbf{The VeNRA Neuro-Symbolic Paradigm.} \textit{(Top)} Standard RAG suffers from ``Stochastic Inaccuracy'' due to probabilistic arithmetic and vector conflation. \textit{(Bottom)} VeNRA parses PDFs into a Universal Fact Ledger (UFL). A Query Navigator retrieves specific variables via a Lexical Gate. An Architect LLM generates a deterministic Python trace. Crucially, the $3$B Sentinel model forensically audits the logic trace before surfacing the answer.}
\label{fig:architecture_overview}
\end{figure*}

Retrieval-Augmented Generation (RAG) has become the standard for open-domain Question Answering (QA) and is widely regarded as the leading paradigm for grounding AI outputs in source material \cite{dadopoulos2025metadata}. However, its deployment in high-stakes, deterministic domains, specifically financial auditing, remains stalled by a phenomenon we term \textit{Stochastic Inaccuracy}. In these environments, an accuracy rate of 99\% yields 0\% operational trust if the 1\% error constitutes a sign-convention inversion on a balance sheet or a temporal hallucination in a compliance report. In fact, as PHANTOM demonstrates, financial QA demands exact numerical precision; even minor errors can critically undermine trust \cite{ji2025phantom}.

This friction arises from two fundamental architectural mismatches. First, dense vector retrieval models rely on \textit{distributional semantics}. Consequently, they frequently map mathematically opposite terms (e.g., ``Net Income'' vs. ``Net Loss'') to proximate vectors due to their identical linguistic contexts, leading to catastrophic retrieval conflation. Basic dense retrieval often fails to fetch the correct evidence in financial documents \cite{dadopoulos2025metadata}, and general-purpose embeddings struggle with finance-specific terms (e.g., ``short'' vs. ``long'') \cite{anderson2024greenback}, consistent with our observation of conflating semantically opposite terms. Second, Large Language Models (LLMs) operate as probabilistic next-token predictors, not arithmetic engines. When tasked with financial reasoning, they simulate the \textit{syntax} of calculation without preserving its mathematical invariants, resulting in plausible-sounding but factually baseless figures \cite{wang2025finlora}. As noted in recent work, LLMs inherently struggle with tasks requiring exact computation \cite{chang2025thor}, producing outputs that sound plausible but violate arithmetic invariants.

Current methodologies to mitigate these hallucinations rely on two fragile pillars: parametric scaling (utilizing frontier models with higher parameters) and generative evaluation benchmarks (e.g., HaluEval). Both are insufficient for production financial systems. Frontier models introduce prohibitive latency ($>2000$ms), rendering them unusable for real-time guardrails. In practice, GPT-4 with retrieval often fails on finance questions (81\% error/refusal rate) \cite{islam2023financebench}, and feeding longer contexts only yields marginal gains with unacceptable latency \cite{islam2023financebench}. More critically, current evaluation benchmarks rely almost exclusively on \textit{Generative Noise}, prompting an LLM to invent an error. For example, the HaluEval benchmark generates false answers via an LLM perturbation framework \cite{li2023halueval}, producing semantically obvious, structurally lazy lies (``Strawman Hallucinations''). Real-world financial agents do not fail by inventing fictitious corporate acquisitions; they fail via \textit{Ecological Errors}: precise mechanical failures such as selecting an adjacent temporal column in a table (Numeric Neighbor Traps) or executing correct Python logic on incorrect variable extractions (Logic Code Lies). Indeed, financial hallucinations often follow specific patterns (numerical miscalculations, temporal shifts, etc.) \cite{tan2025fred}, which generic benchmarks do not capture.

In this paper, we argue that reliable financial AI cannot be achieved purely through parametric scaling or generative evaluation. Instead, it requires \textit{Neuro-Symbolic Cognitive Offloading} and \textit{Forensic Auditing}. We introduce the \textbf{VeNRA\footnote{The code is available at Github: https://github.com/pagand/VeNRA} (Verifiable Numerical Reasoning Agent)} framework, an end-to-end ecosystem designed to prevent, simulate, and detect financial hallucinations. To \emph{mitigate} hallucinations, VeNRA replaces unstructured text retrieval with the UFL. By strictly extracting text into typed variables backed by \textit{Double-Lock Grounding} (requiring both character-offset alignment and semantic schema validation), and gating semantic retrieval with deterministic lexical constraints, we relegate the LLM entirely to the role of a deterministic Python code architect. To \emph{detect} slipped hallucinations, we introduce the \textbf{VeNRA Sentinel}, a locally deployable $3$-B parameter SLM capable of auditing complex mathematical traces in $<50$ms, zero-shot, zero test-time compute. To train the Sentinel, we pioneer \textbf{Adversarial Simulation}, rejecting generative noise in favor of \textbf{VeNRA-Data}: a 10,000-sample dataset constructed by programmatically sabotaging golden financial records to recreate the exact mechanical failures of production RAG systems.

To \textbf{optimize} the Sentinel for zero-latency inference, we adopt a ``Label-First'' generation paradigm. We identify the phenomenon of \textit{Loss Dilution} in Reverse-Chain-of-Thought training, where the gradient signal of the verdict token is overwhelmed by the subsequent reasoning tokens. To solve this, and to prevent Vocabulary Out-Of-Memory (OOM) errors caused by massive differential weighting, we present a novel \textbf{Micro-Chunking Trainer}. This algorithm bounds VRAM usage to $\mathcal{O}(c \cdot |V|)$ rather than sequence length, enabling stable training of highly discriminative judges on consumer hardware. In summary, our main contributions are:
\begin{itemize}
\item \textbf{Neuro-Symbolic Architecture:} The UFL with Double-Lock Grounding and Lexical Pre-Filtering, virtually eliminating extraction hallucination and distributional vector conflation.
\item \textbf{Adversarial Simulation (VeNRA-Data)\footnote{ https://huggingface.co/datasets/pagand/venra}:} A novel methodology for generating rigorous hallucination benchmarks via deterministic sabotage (e.g., Logic Code Lies), evaluated via a strict Hybrid-Family splitting protocol to prevent dataset leakage.
\item \textbf{Forensic Auditing Systems Engineering\footnote{https://huggingface.co/spaces/pagand/VeNRA\_halDet}:} The VeNRA Sentinel, a 3B parameter judge achieving low latency verification. We contribute the Micro-Chunking Trainer, mathematically solving the gradient dilution problem in Reverse-CoT architectures.
\end{itemize}

\section{Background and Related Work}
\label{sec:background}
Neural RAG has become the standard architecture for knowledge-intensive tasks \cite{lewis2020rag}. However, its application in high-stakes, deterministic domains such as financial auditing reveals systematic mismatches between probabilistic language modeling and strict numerical reasoning. While existing literature addresses financial QA benchmarks, hallucination detection, and neuro-symbolic execution in isolation, VeNRA integrates these streams to address the specific failure modes of production financial systems.

\subsection{Failures of Dense Retrieval in Finance}
Classical RAG architectures rely on dense vector retrieval (e.g., DPR, ColBERT) to map queries to evidence \cite{karpukhin2020dense, khattab2020colbert}. While effective for encyclopedic semantic matching, these models suffer from \textit{distributional conflation} in financial contexts. Embeddings optimized for semantic similarity frequently cluster mathematically opposite terms (e.g., ``A owes B'' vs. ``B owes A'') due to identical linguistic contexts \cite{thakur2021beir}. Furthermore, recent studies on \textit{FinanceBench} demonstrate that generic embedding models struggle with the precise temporal and entity alignment required for 10-K filings, where the answer often hinges on distinguishing ``2022'' from ``2021'' within dense tabular structures \cite{islam2023financebench}. VeNRA addresses this by shifting from pure semantic retrieval to a hybrid Lexical-Semantic Gate, prioritizing deterministic token overlap for discriminative financial variables.
\subsection{Numerical Reasoning and Neuro-Symbolic Agents}
Large Language Models (LLMs) act as probabilistic next-token predictors, often failing to maintain arithmetic invariants during text generation \cite{patel2021nlp}. To mitigate this, recent work has pivoted toward neuro-symbolic methods. \textit{Chain-of-Thought} (CoT) prompting \cite{wei2022chain} improves reasoning but remains prone to calculation errors. \textit{Program-Aided Language Models} (PAL) and \textit{Program of Thoughts} (PoT) advance this by delegating computation to an external Python interpreter \cite{gao2023pal, chen2022program}.

However, benchmarks like \textit{FinQA} \cite{chen2021finqa} and \textit{TAT-QA} \cite{zhu2021tat} reveal that even code-generating models struggle with \textit{grounding}; frequently executing correct logic on hallucinated input variables. VeNRA extends the PAL paradigm by enforcing a stricter separation of concerns: the LLM acts solely as a \textit{Code Architect} operating over a strictly typed \textit{UFL}, preventing the injection of unsupported scalars into the computation graph.

\subsection{Hallucination Benchmarks: Generative vs. Ecological}
The evaluation of hallucinations has largely relied on datasets generated via LLM perturbation. Benchmarks like \textit{HaluEval} \cite{li2023halueval} and \textit{TruthfulQA} \cite{lin2021truthfulqa} use strong models to rewrite factual sentences into plausible falsehoods. While scalable, this method produces ``Semantic Noise'', hallucinations that are often linguistically distinct from the source. In contrast, financial hallucinations are often \textit{mechanical}: a column shift in a table or a unit swap (e.g., millions to billions). These ``Ecological Errors'' are underrepresented in generative benchmarks. New benchmarks like FinTextQA and FINDER have been proposed to capture more realistic, expert-level financial queries \cite{dadopoulos2025metadata, chen2024fintextqa, choi2025finder}, yet they focus primarily on QA accuracy rather than auditing execution traces. VeNRA-Data diverges from prior work by using \textit{Adversarial Simulation}, deterministic programmatic sabotage of execution traces, rather than generative rewriting, aligning closer to Red Teaming methodologies \cite{ganguli2022red} than traditional QA dataset creation.

\subsection{LLM-as-a-Judge and Low-Latency Auditing}
Using LLMs as judges to evaluate RAG outputs is a burgeoning field \cite{zheng2024judging}. However, most ``Judge'' architectures rely on frontier models (e.g., GPT-4) which incur multi-second latencies, associated cost, privacy concern, inconsistency, prompt sensitivity, bias, etc rendering them unsuitable for real-time guardrails \cite{kim2023prometheus}. Furthermore, standard instruction tuning for these judges optimizes for fluent rationales rather than calibrated classification, leading to ``sycophancy'' \cite{sharma2023sycophancy}.
VeNRA’s Sentinel model contributes to the growing literature on efficient \textit{Small Language Models} (SLMs) \cite{gunasekar2023textbooks} and single-pass classification. By utilizing a ``Label-First'' paradigm to tether gradients and a novel Micro-Chunking loss to manage vocabulary memory, we demonstrate that a 3B parameter model can achieve frontier-level auditing performance with single token test-time budget, overcoming the latency-accuracy trade-off inherent in prior LLM-as-a-Judge work.
 
\subsection{The Long-Context Fallacy}
Finally, VeNRA challenges the assumption that expanding context windows (e.g., to 128k tokens) solves retrieval challenges. Consistent with the ``lost-in-the-middle'' phenomenon, LLMs often overlook information buried deep in long contexts \cite{liu2023lost}, which is critical for footnotes in 10-K filings. VeNRA’s \textit{Double-Lock Grounding} and \textit{Prompt Bookending} strategies explicitly mitigate this by structuring the context via the UFL rather than relying on the model’s raw attention span over unstructured text.

\section{The VeNRA Architecture: Grounded Neuro-Symbolic RAG}
\label{sec:architecture}

Standard RAG architectures model high-stakes QA as a purely probabilistic sequence generation task. In deterministic domains such as finance, this paradigm induces \textit{stochastic inaccuracy}: LLMs simulate the syntax of mathematics rather than computing it, and dense vector embeddings frequently conflate mathematically opposite but contextually proximate terms (e.g., ``Deferred Revenue'' vs. ``Realized Revenue''). 

To mitigate these fundamental flaws, we introduce VeNRA to shift the RAG paradigm from ``Retrieving Text'' to ``Retrieving Variables''. Rather than forcing a single LLM to simultaneously act as a semantic search engine, an unstructured parser, and an arithmetic calculator, VeNRA strictly decouples these cognitive burdens. Unstructured documents are deterministically parsed into a strongly typed UFL, mathematically grounded, and subsequently queried by an LLM constrained strictly to Python code generation.

\subsection{Ingestion via Universal Fact Ledger}
\label{subsec:ingestion}

The UFL serves as the foundational data contract for the VeNRA ecosystem, acting simultaneously as a computational table for arithmetic and a relational graph for semantic traversal. To populate the UFL from unstructured 10-K filings without succumbing to ``Context Rot'' \cite{hong2025context}, the loss of structural lineage and coreference, we employ a deterministic ingestion pipeline.

Standard recursive chunking mechanisms frequently sever cross-boundary coreferences (e.g., a table row referencing ``See Note 12'', where the note text resides in the subsequent chunk). To preserve localized context, we developed a newline-preference trailing-buffer algorithm. 
Let $C_i$ be the $i$-th chunk of text, bounded by a character threshold $\tau = 3000$. Rather than splitting at an arbitrary character index, the algorithm splits at the most recent newline character. For every generated text block $C_i$, a trailing buffer $b_{i-1}$ consisting of the final $k=300$ characters of the preceding block $C_{i-1}$ is injected as a prefix: 
\begin{equation}
    C'_i = [\text{Previous Context: } b_{i-1}] \oplus C_i
\end{equation}
Crucially, table structures also update this rolling buffer. Because financial table headers carry essential scalar context (e.g., ``in millions''), propagating the table's raw tail into the subsequent prose chunk allows the extraction model to successfully resolve anaphoric scalar references.

Generative LLMs exhibit severe unreliability when extracting 2D grid data \cite{zhu2021tat}. Consequently, VeNRA bypasses LLM extraction for tables entirely, employing deterministic Pandas routines. Raw markdown tables are parsed using a hierarchical disambiguation stack. Indentation markers (e.g., \texttt{\&nbsp;}) are computed to generate fully qualified schema paths (e.g., \textit{Assets > Current Assets > Inventory}), preventing the collapse of distinct sub-categories into ambiguous metrics like ``Other.'' Furthermore, row-level overrides dynamically intercept global column scale factors; metrics containing substrings such as ``per share'' enforce a $1.0$ multiplier and a unit classification of \texttt{USD/Share}, ensuring mathematical invariants are preserved.

\subsection{Double-Lock Grounding and Post-Hoc Alignment}
\label{subsec:grounding}

The primary vulnerability of JSON-constrained LLM extraction is the generation of unsupported numerical values. To solve the ``Generator vs. Extractor'' dilemma, we force the generative model to provide a \texttt{grounding\_quote} ($q$), the verbatim substring from the source chunk ($C$) that justifies the extracted scalar ($v$). Every extracted fact undergoes a rigorous \textit{Double-Lock Grounding} protocol.

\textbf{Lock 1: Mechanical Quote Grounding.} 
The algorithm attempts to physically map the quote $q$ to exact character offsets in $C$. Standard normalized sequence matching algorithms (e.g., \texttt{difflib}) over entire text chunks were discarded; we observed that sequence matching against a large denominator (e.g., a 3000-character chunk) asymptotically approaches 0, rendering it empirically useless for isolated numeric verification. Instead, we use a three-tier fallback:
\begin{itemize}
    \item \textit{Tier 1 (Exact Match):} An $\mathcal{O}(n)$ character-exact substring search, followed by a whitespace-normalized pass (collapsing $\setminus s+$ to a single space) to account for LLM-induced spacing artifacts.
    \item \textit{Tier 2 (Partial Match):} The aligner extracts the longest contiguous numeric token $q_{num}$ from $q$ (e.g., ``615'' from ``\$615 million'') and searches for it verbatim, verifying the scalar value independently of surrounding text hallucination.
    \item \textit{Tier 3 (Fuzzy Sliding-Window):} A sliding window $W$ of tokens is passed over $C$. We deliberately avoid Jaccard Similarity, which heavily penalizes the intersection due to the large union of the document window ($|W| \gg |q|$). Instead, we compute a fast token intersection based on Recall: $R = \frac{|q \cap W|}{|q|}$. If $R > 0.55$, the window is refined with a localized character-matching algorithm.
\end{itemize}
If all three tiers fail, the fact is stamped \texttt{UNALIGNED}, confidence is set to $0.0$, and the scalar is restricted from downstream execution.

\textbf{Lock 2: Semantic Metric Grounding.} 
Mechanical grounding prevents invented numbers, but it does not prevent ``Phantom Metrics'', a hallucination where an LLM assigns a verifiably real number to a fabricated metric name. Let $M$ be the set of core tokens in the generated metric name, excluding general financial stop-words (e.g., ``net'', ``total'', ``ratio''). Let $T_C$ be the set of tokens in the source chunk. The semantic overlap score is defined as $s = \frac{|M \cap T_C|}{|M|}$. If $s < \tau_{sem}$ (where $\tau_{sem} = 0.30$), the fact is rejected as a semantic hallucination, successfully preventing the model from assigning orphaned numbers to hallucinatory schemas.

\subsection{Schema-Driven Cognitive Offloading}
\label{subsec:schema_offload}

A persistent challenge in applied LLM engineering is ``prompt overfitting'', attempting to handle infinite edge cases via increasingly complex system instructions. In finance, edge cases include dynamic formulas (e.g., \textit{``floating rate equal to SOFR plus 2.25\%''}) and covenants (e.g., \textit{``leverage ratio not exceeding 3.50''}). A naive prompt instructing the LLM to differentiate these leads to downstream mathematical catastrophes when the LLM extracts ``2.25'' as a fixed interest rate.

VeNRA introduces \textit{Schema-Driven Cognitive Offloading}. Rather than relying on conditional logic in the prompt, we leverage the LLM's inherent pre-training in taxonomy by introducing a Pydantic classification enum: \texttt{fact\_type $\in$ \{ACTUAL, LIMIT, FORMULA\}}.  Once the model classifies the fact via constrained decoding, deterministic Python post-processing handles the operational logic. For \texttt{FORMULA} types, the system forcefully nullifies the float (\texttt{num\_value = None}). This creates a critical safety boundary: when the Code Agent later queries the UFL, it encounters a \texttt{None} type and reads the formula string in the \texttt{text\_nuance} metadata, preventing it from executing arithmetic on a partial floating rate. For \texttt{LIMIT} types, the system automatically appends the suffix \texttt{[Limit]} to the metric name, preserving the strict ontological boundary between an actual performance metric and a legal constraint.

\subsection{Hybrid Lexical-Semantic Retrieval}
\label{subsec:retrieval}

Once the UFL is populated, the system must retrieve the precise variables required to satisfy a user query. Pure semantic vector search is fundamentally ill-suited for financial retrieval because dense embedding models (e.g., \texttt{bge-m3}) rely on distributional semantics. Consequently, they frequently group mathematically opposite terms by contextual proximity. A query for ``Net Sales'' will consistently retrieve chunks discussing ``Net Income'' with high cosine similarity, leading to cascading arithmetic failures downstream.

To overcome this, VeNRA introduces a \textit{Hybrid Lexical-Semantic Retriever}, governed by an LLM-based Query Navigator \cite{agand2024dmfuser} and a deterministic Lexical Gate.

Upon receiving a query, a Navigator SLM evaluates the user intent against a dynamically injected schema summary, which maps canonical entity IDs and high-frequency metric names. The Navigator outputs a bipartite \texttt{RetrievalPlan} containing: (1) a structured UFL filter (e.g., Entity ID, Year, Metric Keywords, and Nuance conditions), and (2) a ``Vector Hypothesis'', an idealized text snippet theoretically containing the answer.

To prevent distributional semantic conflation, all vector retrieval candidates (and fuzzy UFL metric matches) must pass a deterministic Lexical Gate before entering the context window. 

Let $Q$ be the multiset of normalized tokens in the query, and $C$ be the multiset of tokens in the candidate chunk. Standard Jaccard Similarity ($J = \frac{|Q \cap C|}{|Q \cup C|}$) is strictly avoided; because financial text chunks are highly verbose ($|C| \gg |Q|$), the massive union denominator forces $J$ toward zero even for perfect semantic matches. Instead, we compute token intersection Recall:
\begin{equation}
    R_{lex} = \frac{\sum_{t \in \text{set}(Q)} \min(\#(t, Q), \#(t, C))}{|Q|}
\end{equation}
where $\#(t, X)$ denotes the multiplicity of token $t$ in multiset $X$. If $R_{lex} < \tau_{lex}$ (where $\tau_{lex} = 0.30$), the candidate is dropped. 

Crucially, two pre-processing steps guarantee the gate's efficacy:
\begin{enumerate}
    \item \textbf{Domain-Specific Stop-Word Filtering:} Standard financial modifiers (e.g., ``net'', ``total'', ``per'') are strictly classified as non-discriminative stop-words and removed from $Q$. If ``net'' were retained, a query for ``Net Income'' ($Q = \{\text{net}, \text{income}\}$) evaluated against a candidate discussing ``Net Sales'' ($C = \{\text{net}, \text{sales}\}$) would yield $R_{lex} = 0.50$, falsely passing the threshold. Filtering ``net'' isolates the discriminative tokens, yielding $R_{lex} = 0.0$ and successfully blocking the semantic hallucination.
    \item \textbf{Prefix Stripping:} The trailing context buffer ($[\text{Previous Context: } b_{i-1}]$) injected during the chunking phase (Section \ref{subsec:ingestion}) is forcefully stripped from $C$ prior to tokenization. This prevents cross-chunk token contamination, ensuring that an irrelevant candidate does not pass the gate simply because its injected prefix shares tokens with the query.
\end{enumerate}

\textbf{Relational Graph Expansion.} 
Because the UFL acts as a relational graph, retrieval operates along three concurrent expansion pathways:
\begin{enumerate}
    \item \textit{Structured Expansion:} The UFL DataFrame is filtered sequentially. If a \texttt{nuance\_focus} is generated by the Navigator (e.g., ``Restated''), it acts as a boolean AND-gate over the \texttt{text\_nuance} column, deterministically isolating adjusted figures.
    \item \textit{Entity Pivoting:} If retrieved UFL rows contain a \texttt{related\_entity\_id} (e.g., a supplier), the system pivots and queries the vector database for chunks mentioning that specific entity, using a relaxed lexical threshold ($\tau_{lex} = 0.20$) to account for sparse entity mentions in prose.
    \item \textit{Frequency-Based Chunking:} To ensure the reasoning agent has access to the qualitative text (the ``data hub'') that generated the numbers, the system isolates the most frequent \texttt{source\_chunk\_id}s from the retrieved UFL rows and forcefully includes their raw text chunks in the context window.
\end{enumerate}

\subsection{Program-Aided Execution and Context Assembly}
\label{subsec:execution}

The final architectural phase extends the Program-Aided Language Models (PAL) paradigm \cite{gao2023pal}. The LLM is stripped of its role as a probabilistic calculator and relegated to the role of a deterministic software architect.

\textbf{Context Assembly and Priority Scoring.} 
To prevent ``Context Rot'', where an LLM loses attention over an expansive context window, the Context Assembler deduplicates all retrieved UFL rows and text chunks. Chunks are ranked via a priority scoring heuristic: a chunk $c$ receives $+5$ points if it is physically linked to a retrieved UFL row (via \texttt{source\_chunk\_id}), and $+1$ point for every BM25 keyword it contains. The context is strictly truncated to the top $k=5$ chunks and presented as a fused Markdown representation containing both the structured UFL table and the unstructured raw text.

VeNRA executes reasoning via a dual-pass neuro-symbolic flow:
\begin{itemize}
    \item \textbf{Pass 1 (The Architect):} A reasoning model evaluates the assembled context. It generates a step-by-step logic plan and writes deterministic Python code to compute the answer. The code is executed in an isolated local Python subprocess. If a formula constraint was previously offloaded (Section \ref{subsec:schema_offload}) and the UFL \texttt{num\_value} is \texttt{None}, the code naturally halts or explicitly requests the missing variable from the text, preventing blind arithmetic.
    \item \textbf{Pass 2 (The Synthesizer):} A low-latency SLM ingests the mathematical output from the Python subprocess and formulates the final natural language response, ensuring every stated number is cited directly to a UFL \texttt{row\_id} or Chunk ID.
\end{itemize}

Despite rigorous ingestion grounding, optical character recognition (OCR), failure in UFL underlying LLM, and PDF parsing anomalies inevitably occur. VeNRA implements a programmatic fault-tolerance mechanism. The Architect agent is explicitly instructed via its system prompt to cross-reference the structured UFL rows against the provided raw text chunks. If the agent detects that the UFL data is malformed (e.g., an unaligned temporal column) or mathematically contradicts the raw text, it programmatically discards the UFL variable. Instead, the Python code is written to extract and compute the values directly from the provided source text string, guaranteeing system resilience.

\section{VeNRA-Data: Adversarial Simulation of RAG Failures}
\label{sec:dataset}

To effectively train a low-latency Sentinel model to audit the outputs of the VeNRA architecture, we require a dataset that accurately reflects the failure modes of financial RAG systems. We introduce \textbf{VeNRA-Data}, a heavily curated dataset comprising contrastive pairs of grounded and ungrounded financial reasoning traces.

To prevent the classification model from becoming overly dependent on contextual presence, a common flaw where models flag true financial axioms as hallucinations simply because they are omitted from the immediate text, we avoid the binary True/False paradigm. Instead, VeNRA-Data adopts a strict tripartite taxonomy: \texttt{SUPPORTED} (backed by text or valid computation), \texttt{UNFOUNDED} (contradicts evidence, uses ungrounded inputs, or commits logic errors), and \texttt{GENERAL} (axiomatic financial truths independent of the provided context).

\subsection{The Flaw in Generative Hallucinations}
\label{subsec:flaw_generative}

Current hallucination detection benchmarks (e.g., HaluEval) primarily rely on LLM-generated semantic noise. Dataset creators typically prompt a frontier model to intentionally rewrite a factual sentence into a hallucination. This methodology fundamentally lacks \textit{ecological validity}. Generative hallucinations tend to be linguistically obvious, structurally lazy, and linguistically distinct from the source text. 

Financial RAG systems in production do not fail by inventing linguistically anomalous sentences; they fail via precise mechanical, calculative, and mapping errors. An LLM extracting data from a 10-K is far more likely to select an adjacent temporal column (e.g., $2022$ instead of $2023$) or divide by a hallucinated denominator than it is to invent a fictitious corporate acquisition. 

To overcome the limitations of generative noise, we shift the paradigm to \textbf{Adversarial Simulation}. We utilize five high-quality, human-annotated financial QA datasets including FinQA \cite{chen2021finqa}, TAT-QA \cite{zhu2021tat}, FinanceBench \cite{islam2023financebench}, truthfulQA \cite{lin2022truthfulqa}, and PHANTOM \cite{ji2025phantom} as our ``Golden'' baseline. Rather than asking an LLM to rewrite them, we employ a deterministic \textit{Saboteur Engine} to systematically mutate specific informational pathways, algorithmically recreating the exact cognitive failures observed in production architectures.

\subsection{The Saboteur Engine: Programmatic Injections}
\label{subsec:saboteur}

The Saboteur Engine consists of modular, deterministic perturbation algorithms. For a given Golden record containing a Query ($Q$), Context ($C$), execution Trace ($T$), and Target Sentence ($S$), the engine attempts to generate an \texttt{UNFOUNDED} ``Hard Negative'' clone via one of four specific attack vectors.

\textbf{1. The Logic Code Lie.} 
This vector targets records containing arithmetic traces (e.g., FinQA). It simulates a scenario where the generated Python code is syntactically and mathematically valid, but the input variables hallucinated by the agent are fundamentally incorrect. 
Let $V_T$ be the set of numeric scalars present in the execution trace $T$, and $V_C$ be the set of numeric scalars present in the context $C$. The engine isolates a target variable $v \in V_T$ (excluding structural indices such as $0$ or $1$). It then samples a distractor scalar $d \in \{V_C \setminus V_T\}$. The trace is mutated via functional substitution: $T' = T[v \leftarrow d]$. 

Crucially, the Saboteur then executes the hallucinated trace $T'$ in a secure sandbox to produce a new resultant scalar $R'$. The target sentence is updated to reflect $R'$ ($S \leftarrow S'$). The resulting record presents a profound challenge to the Judge model: the sentence $S'$ perfectly matches the output of the trace $T'$, but the trace relies on an input $d$ that misaligns with the semantic text $C$.

\textbf{2. The Numeric Neighbor Trap.}
This vector targets tabular extraction failures, simulating the ``column/row shift'' errors endemic to PDF optical character recognition (OCR) and LLM grid parsing. Let the financial table be represented as a 2D matrix $M$. The engine locates the ground-truth scalar within the matrix at coordinates $(i, j)$ such that $M_{i,j} \in S$. The algorithm deterministically shifts the coordinate space to select a neighboring cell $M'$, where $M' \in \{M_{i\pm1, j}, M_{i, j\pm1}\}$. 

By substituting $M'$ into $S$, the engine simulates two highly prevalent RAG failures: a \textit{Temporal Slip} (a horizontal column shift, retrieving the wrong fiscal year) or a \textit{Metric Slip} (a vertical row shift, retrieving an adjacent accounting category).

\textbf{3. Irrelevancy and Time Warps.}
This vector simulates staleness and retrieval failure. 
\begin{itemize}
    \item \textit{The Time Warp:} If the user query $Q$ contains a specific temporal bounding condition (e.g., the year $2020$), the Saboteur mutates the query $Q \leftarrow Q_{2021}$. The context $C$ and the sentence $S$ remain entirely unchanged. This trains the model to detect scenarios where the provided answer is factually supported by the text, but fails to satisfy the temporal constraints of the user's prompt.
    \item \textit{Context Swap (Non-Existence):} The engine retains $Q$ and $S$, but replaces the context $C$ entirely with text chunks sampled from a completely disjoint financial document. This creates a ``non-existence'' hallucination, forcing the Sentinel model to detect when a generated answer is entirely ungrounded by the provided retrieval context.
\end{itemize}

\textbf{4. Semantic and Scale Drift.}
Finally, the engine applies deterministic regex-bounded substitutions to simulate unit and entity mapping failures. Using a predefined domain ontology, scale tokens are inverted (e.g., ``millions'' $\leftrightarrow$ ``billions''), and capitalized Named Entities are swapped with disjoint entities found elsewhere in the text space (e.g., misattributing a subsidiary's revenue to the parent corporation).

\subsection{The Teacher-Auditor Protocol and Noise Injection}
\label{subsec:teacher_auditor}

While programmatic sabotage deterministically generates hard negatives, it is susceptible to ``Accidental Truths'', scenarios where an injected scalar coincidentally matches a valid contextual fact, or where the original ``Golden'' record itself contains human annotation errors (e.g., ``Dirty Gold''). 

To guarantee the pristine quality of the training signals, VeNRA-Data employs a multi-tiered \textit{Teacher-Auditor Protocol}. The Auditor is tasked with independently evaluating the sabotaged tuple $(Q, C, T', S')$ to verify if it is genuinely unfounded.

A critical algorithmic hurdle in automated auditing is the decoupling of cause and effect. In a ``Logic Code Lie'' (Section \ref{subsec:saboteur}), the Saboteur mutates an input variable ($v \leftarrow d$). However, the Teacher model frequently flags the \textit{output} scalar (the final computed answer in $S'$) as the error span, rather than the injected input $d$. A naive string-matching validation script falsely rejects these valid sabotages because the cause ($d$) does not equal the effect ($S'$).

We resolve this via a \textit{Dual-Path Validation} algorithm. Let $E_{teacher}$ be the error span detected by the Teacher, and $\Theta_{teacher}$ be the internal reasoning trace (the \texttt{<think>} block). A sabotage injection $d$ is validated if it satisfies either of the following conditions:
\begin{enumerate}
    \item \textit{Effect Match:} The fuzzy token set ratio between $d$ and $E_{teacher}$ exceeds a threshold.
    \item \textit{Cause Match:} The injected scalar $d$ is explicitly present within the Teacher's causal reasoning block $\Theta_{teacher}$.
\end{enumerate}

To train the \texttt{GENERAL} class, we utilize TruthfulQA's finance and economics splits. Initially, these records contain an empty context ($C = \emptyset$). However, training an autoregressive model on this representation induces a spurious correlation: the model learns the structural shortcut $C = \emptyset \implies \texttt{GENERAL}$, bypassing semantic reasoning entirely. To force true logic-awareness, we implement \textit{Axiomatic Noise Injection}. For every general axiom, we sample a subset of completely irrelevant financial text chunks from a distractor pool ($C_{distractor}$). The model must learn to actively ignore the distractor context and rely on its parametric memory when the trace signals \texttt{GENERAL\_KNOWLEDGE}, preventing structural overfitting.   For highly ambiguous samples where the automated validation yields low confidence, we utilize an AI Proxy Refiner (More on Appendix.\ref{app:hitl}).

\subsection{Contrast Pairs and Hybrid-Family Splitting}
\label{subsec:hybrid_splitting}

The final architectural requirement for VeNRA-Data is the prevention of dataset leakage and the establishment of a causal evaluation framework. 

Standard randomized stratification (e.g., an 80/10/10 train/val/test split) over a dataset containing cloned records guarantees severe data leakage. If a Golden record (the Parent) is assigned to the training set, and its sabotaged clone (the Child) is assigned to the validation set, the model evaluates on a context it has already memorized, inflating performance metrics.

To eliminate leakage and maximize the learning signal, we implement \textbf{Hybrid-Family Splitting}. 
Let a Family $F_i$ be defined as a set containing a golden Parent record and its corresponding sabotaged Children: $F_i = \{S_{parent}, S'_{child\_1}, \dots\}$. Families are derived programmatically via a lineage heuristic (\texttt{family\_id}). 

The dataset is partitioned not by individual records, but by entire Families. Families are grouped into four distinct semantic buckets:
\begin{enumerate}
    \item \textbf{Sabotage Pairs:} Families containing both \texttt{SUPPORTED} and \texttt{UNFOUNDED} siblings.
    \item \textbf{Natural Failures:} Families containing organically \texttt{UNFOUNDED} records (e.g., original FinanceBench failures) with no supported sibling.
    \item \textbf{Natural Supported:} Families comprising only golden records not selected for sabotage.
    \item \textbf{Axioms:} Families representing \texttt{GENERAL} truths.
\end{enumerate}

A stratified split is performed across these buckets, ensuring exactly 10\% of the \textit{Families} from each bucket are allocated to the Validation and Test sets respectively. 

This strict topological isolation provides a dual benefit: it mathematically guarantees zero contextual leakage between splits, and it forces the model to learn the \textit{delta}. By encountering identical contexts and queries during training, where one answer is mathematically sound and the other contains a subtle logic lie, the model's loss landscape is optimized to detect minute reasoning derivations rather than broad linguistic artifacts. Furthermore, preserving these pairs in the evaluation sets enables the calculation of a strict ``Flip Rate'' (Sensitivity), measuring the model's ability to correctly invert its prediction when presented with a minimal adversarial perturbation.

\section{VeNRA Sentinel: Forensic Auditing under Latency Constraints}
\label{sec:sentinel}

To deploy VeNRA in a production financial environment, the generated answers must be audited in real-time. Frontier LLM-as-a-Judge models introduce prohibitive latency ($\approx 2$-$4$ seconds) and excessive operational costs. To solve this, we introduce the \textbf{VeNRA Sentinel}, a specialized $3$-B parameter Small Language Model (SLM) based on \texttt{Qwen-2.5-Coder-3B-Instruct}\footnote{ https://huggingface.co/pagand/venra}. 

The Sentinel is trained as a ``White-Box'' hallucination detector. It ingests the User Query ($Q$), the Text Context ($C$), the Python Execution Trace ($T$), and the Target Sentence ($S$), outputting a strict tripartite verdict: \texttt{SUPPORTED}, \texttt{UNFOUNDED}, or \texttt{GENERAL}. To satisfy a strict inference latency budget while maintaining reasoning capabilities, we had to fundamentally alter standard instruction-tuning paradigms.

\subsection{The System 1.5 Paradigm and Token Orthogonality}
\label{subsec:salsa}

Standard Chain-of-Thought (CoT) prompting dictates that reasoning must precede the final answer (e.g., \texttt{[Analysis] $\rightarrow$ [Label]}). While this allows the model to utilize computational ``test time'' to arrive at a conclusion, autoregressive generation of a 150-token analysis incurs $\approx 1500$ms of latency on standard consumer hardware, violating our latency constraints.

To satisfy latency constrain, we inspired from SALSA (Single-pass Autoregressive LLM Structured Classification) \cite{berdichevsky2025salsa} framework, formatting the training targets as Reverse-CoT:\\ \texttt{Label: [Verdict] \textbackslash n Analysis: [Reasoning]}. 

In an autoregressive framework, tokens generated \textit{after} the label cannot mathematically influence the probability distribution of the label itself during a single forward pass. However, training the model to predict the label and immediately explain it acts as a ``Gradient Tether.'' It forces the hidden states of the \texttt{Label} token to become incredibly dense, requiring it to contain not just the binary classification decision, but the precise latent ``seed'' of the upcoming forensic explanation. During production inference, we execute a single forward pass, extract the unnormalized logits for the label tokens, and programmatically halt generation, achieving latency verification. If a query is flagged as \texttt{UNFOUNDED}, generation can optionally resume to provide the debug analysis to the end-user.

To extract highly calibrated logits at the very first decoding step, the target tokens must be semantically and mathematically distinct. Naive binary labels (e.g., \texttt{True} / \texttt{False} or \texttt{SUPPORTED} / \texttt{UNFOUNDED}) suffer from severe semantic conflation in the model's pre-training data. 

To select optimal classification tokens, we extracted the input embedding matrix $E \in \mathbb{R}^{|V| \times d}$ from the base \texttt{Qwen} model and computed pairwise cosine similarities:
\begin{equation}
    sim(e_i, e_j) = \frac{e_i \cdot e_j}{\|e_i\| \|e_j\|}
\end{equation}
We discovered that standard labels like \texttt{True} / \texttt{False} exhibited a high cosine similarity ($sim \approx 0.68$), meaning their initial vector representations are highly proximate, forcing the attention layers to work harder to separate them. Through programmatic search, we identified a highly orthogonal set of single-token labels: \texttt{ Found}, \texttt{ Fake} , and \texttt{ General}. This set yielded a maximum pairwise cosine similarity of $< 0.25$, ensuring that the linear classification head operates on easily separable latent vectors.

\subsection{The Effects of Loss Dilution}
\label{subsec:loss_dilution}

Training the Sentinel model using standard trainers (e.g. \texttt{SFTTrainer}) resulted in catastrophic failure. The model rapidly achieved a low validation loss  but exhibited a near-random ``Flip Rate'' accuracy  on the adversarial contrast pairs. The model had learned to be a \textit{Sycophant}: it minimized global loss by defaulting to the majority class (\texttt{Found}) and generating highly fluent, generic financial analyses. 

In a standard autoregressive Cross-Entropy formulation over a sequence of length $N$, the loss is averaged across all active completion tokens:
\begin{equation}
    \mathcal{L}_{batch} = - \frac{1}{N} \sum_{i=1}^{N} \log P(y_i \mid x_{<i})
\end{equation}
In our Reverse-CoT prompt, the sequence consists of $1$ label token and approximately $150$ analysis tokens. Consequently, the gradient contribution of the critical classification decision is profoundly diluted: $\frac{\partial \mathcal{L}}{\partial \theta_{label}} \propto \frac{1}{151} < 0.7\%$. The model is overwhelmingly penalized for grammatical imperfections in its reasoning rather than catastrophic failures in its verdict.

To restore the gradient signal of the verdict, we must apply a large differential weight to the label token. However, implementing a weighted loss via PyTorch's \texttt{CrossEntropyLoss(reduction='none')} on a $3$-B parameter model triggers immediate OOM failures on a single GPU (with <50GB of VRAM)\footnote{For a micro-batch size $B=2$ and sequence length $S=4096$, calculating the unreduced loss instantiates a tensor of size $[B \times S \times |V|]$. At FP32 precision (required for loss stability), this single intermediate tensor consumes $\approx 1.25$ Billion floats ($\approx 5$ GB). Combined with gradients and optimizer states, this vastly exceeds VRAM capacity.}. To solve this, we employed a custom \textbf{Micro-Chunking Trainer}. We bypass the model's internal loss calculation, flattening the shift-logits and shift-labels. We then iteratively slice the sequence into micro-chunks of size $c=512$ entirely within the loss computation loop. 
For a flat sequence length $T = B \times S$, the weighted loss is accumulated safely:
\begin{equation}
    \mathcal{L}_{total} = \frac{\sum_{k=0}^{\lceil T/c \rceil} \sum_{i \in C_k} w_i \mathcal{L}_{CE}(\hat{y}_i, y_i)}{\sum_{k=0}^{\lceil T/c \rceil} \sum_{i \in C_k} w_i}
\end{equation}
This architectural optimization bounds the VRAM consumption of the loss function to $\mathcal{O}(c \times |V|)$ rather than $\mathcal{O}(B \times S \times |V|)$, maintaining exact mathematical equivalence to the full weighted Cross-Entropy.

\subsection{Dynamic Loss Clamping and Differential Weighting}
\label{subsec:loss_clamping}

While the Micro-Chunking Trainer enables weighted loss, multiplying the label gradient by $50\times$ introduces severe topological instability (\textbf{Gradient Seesaw}). 
Imposing a large penalty on a confidently wrong prediction (where $P(\hat{y}) \approx 0$) causes the cross-entropy loss to spike asymptotically\footnote{During early training, we observed massive gradient norm explosions, causing the model to violently oscillate, updating from predicting $100\%$ \texttt{Fake} in one step to $100\%$ \texttt{Found} in the next.}. To stabilize convergence, we introduced \textit{Dynamic Loss Clamping}, explicitly capping the maximum penalization:
\begin{equation}
    \mathcal{L}_{weighted} = \min(w_i \cdot \mathcal{L}_{CE}, \: 5.0 \cdot w_{max})
\end{equation}

Initially, we applied the penalty uniformly across all three classes. This induced a secondary failure mode: the model began drastically over-predicting the minority \texttt{General} class. The model utilized the \texttt{General} token as a ``safe harbor'' to avoid the massive penalization of misclassifying the primary \texttt{Found}/\texttt{Fake} distribution. We resolved this via \textit{Asymmetric Differential Weighting}: assigning a $50\times$ penalty to \texttt{Found}/\texttt{Fake} to force hard binary logic, but reducing the \texttt{General} penalty to $10\times$, preserving axiomatic knowledge retention without inducing class bleed.

To facilitate training at high adapter ranks ($r=96,128$) under these extreme gradient conditions, we utilized 4-bit NF4 Quantization (QLoRA) paired with \textbf{Rank-Stabilized LoRA (rsLoRA)} \cite{kalajdzievski2023rank}. By scaling the adapter activations by $\frac{\alpha}{\sqrt{r}}$ rather than $\frac{\alpha}{r}$, rsLoRA prevents gradient vanishing in deeper layers, enabling stable logic-awareness acquisition.

\subsection{Prompting Physics and Uncertainty Quantification}
\label{subsec:prompt_physics}

Because the base model (\texttt{Qwen-2.5-Coder-3B-Instruct}) possesses a strong inductive bias for code execution, it instinctively attempts to mathematically solve the Python trace provided in the prompt. However, a 3B parameter model lacks the FLOP depth to reliably execute complex 6-digit subtraction in a single forward pass. If it attempts arithmetic simulation, it hallucinates. We counteract this via \textit{Algorithmic De-activation} in the system prompt\footnote{Example prompt: \textit{``Do NOT recalculate the math. Assume the arithmetic evaluates to the Claimed Answer. You must verify the EXTRACTION and LOGIC.''}} to successfully redirects the model's attention heads from arithmetic simulation toward semantic set-intersection and string-matching.

Causal LLMs suffer from the ``Lost in the Middle'' phenomenon in large contexts \cite{an2024make}. To prevent the model from forgetting the verification target after processing 4,000 tokens of 10-K text, we implement \textit{Selective Prompt Repetition}. We  state the Query and Trace at the very top of the prompt, inject the massive context, and strictly repeat the Query and Trace at the bottom boundary immediately preceding the \texttt{Label:} generation token. This aligns with the finding of \cite{leviathan2025prompt} for non-reasoning tasks.

During inference, raw softmax probabilities are deceptive; a model can output $P_{top1} = 0.99$ simply due to softmax temperature, despite high internal confusion \cite{liu2025tokenleveltruth}. Drawing inspiration from recent production safety frameworks such as HaluGate \cite{park2025halloc}, VeNRA implements a Logit Gap uncertainty threshold: $\Delta = P_{top1} - P_{top2}$. If $\Delta < 0.15$, the Sentinel determines the semantic entropy is too high, forcefully overrides the output to \texttt{Uncertain}, and routes the query to a human auditor or a frontier System-2 model.

\section{Experiments and Results}
\label{sec:experiments}

To rigorously evaluate the VeNRA architecture, we decouple the empirical
analysis into two distinct stages.
First, we evaluate the neuro-symbolic extraction and execution pipeline,
isolating the causal impact of the UFL and DualRetriever
against standard dense retrieval.
Next, we evaluate the Sentinel model, demonstrating the efficacy of
Adversarial Simulation and our latency-optimised training dynamics.

\subsection{Experimental Setup}
\label{subsec:setup}

All training runs and GPU-based inference evaluations were conducted on a single
NVIDIA A40 (48\,GB VRAM).
The VeNRA Sentinel was initialised from \texttt{Qwen2.5-Coder-3B-Instruct}.
We allocated high representation capacity ($r{=}128$, $\alpha{=}128$) across
all attention and MLP projection layers to capture the complex extraction and
logic mappings required for forensic trace verification.
The optimiser was \texttt{paged\_adamw\_8bit} with a cosine annealing schedule
decaying to a minimum learning rate of $5{\times}10^{-6}$.

Table~\ref{tab:dataset_stats} details the adversarial corpus distribution.
The 8,320 records are sourced from public financial benchmarks and augmented with a Phantom split of
purely synthetic long-document queries generated to stress-test cross-table retrieval.
Each record anchors to a verified Python execution trace produced by the PAL agent; the \texttt{SUPPORTED} class is an organic correct answer and the \texttt{UNFOUNDED} class is an adversarially sabotaged counterpart.

The dataset is explicitly balanced ($51.3\%$ \texttt{SUPPORTED},
$47.2\%$ \texttt{UNFOUNDED}) to prevent the baseline sycophancy observed when
the model is trained on real-world hallucination corpora with skewed class
ratios.
A deliberate \textsc{General} class ($1.5\%$ of records) acts as an axiomatic
regularisation signal, teaching the Sentinel to invoke parametric memory for
universal financial truths (e.g., that revenue cannot exceed total assets in a
single period) rather than penalising every ungrounded claim.

A sub-sample of adversarial pairs from the
\texttt{pair\_short} and \texttt{pair\_long} pools, with natural failures and
general-knowledge items excluded is used for all comparisons with SoTA models in
Section~\ref{subsec:exp3_sentinel} to ensure that cross-model Flip Rates are measured on structurally identical, paired stimuli.

\begin{table}[ht]
\centering
\caption{VeNRA-Data distribution across splits.
The topological split ensures all sabotage variants of the same source context
reside in the same fold.}
\label{tab:dataset_stats}{%
\begin{tabular}{@{}lcccc@{}}
\toprule
\textbf{Split} &
\textbf{Total} &
\textbf{Supported} &
\textbf{Unfounded} &
\textbf{General} \\
\midrule
Training   & 6,673 & 3,421 & 3,153 &  99 \\
Validation &   835 &   425 &   396 &  14 \\
Testing    &   812 &   420 &   382 &  10 \\
\midrule
\textbf{Total} & \textbf{8,320} & \textbf{4,266} & \textbf{3,931} & \textbf{123} \\
\bottomrule
\end{tabular}}
\end{table}

The Sentinel prompt follows a
five-zone layout within a 4,096-token budget:\\
Zone 1 (\textit{System}) establishes the auditor persona.\\
Zone 2 (\textit{Verification Target}) presents the query, agent trace, and
claimed answer.\\
Zone 3 (\textit{Evidence}) contains the source document context.\\
Zone 4 (\textit{Selective Repetition}) re-presents the query, trace, and claimed answer verbatim \\
Zone 5 (\textit{Audit Algorithm}) specifies a four-step procedure:
(1)~Extraction Check: all numbers and entities in the trace must appear verbatim in the evidence;
(2)~Logic Check: the trace operation and metric selection must match the query
intent;
(3)~Axiom Check: claims not verifiable from evidence but constituting universal financial knowledge yield \texttt{General};
(4)~verdict synthesis: Output \texttt{Found} if both extraction and logic are both supported by the evidence. 

The assistant turn is pre-populated with the literal prefix (\texttt{Label:}). The logit at the last position is read directly over the three
orthogonal class tokens (\texttt{Found}, \texttt{Fake}, \texttt{General}),
and the argmax is taken without any autoregressive continuation.
During training, the completion extends to
\texttt{\{label\}\textbackslash nAnalysis: \{reasoning\}<|im\_end|>},
instantiating the Reverse-Chain-of-Thought paradigm in which the verdict
precedes its justification, the inverse of standard CoT training.

\subsection{Experiment 1: Neuro-Symbolic Pipeline Evaluation}
\label{subsec:exp1_pipeline}

We evaluate the VeNRA pipeline under a 2$\times$2 factorial design crossing the
\textit{retrieval} axis (Baseline Vector Search vs.\ VeNRA DualRetriever with
Navigator rescue) and the \textit{generation} axis (Gemini-1.5-Pro CoT vs.\
VeNRA PAL Code Agent).  All four configurations are evaluated on  held-out
questions drawn from three financial benchmarks: TAT-QA ($35\%$),
FinQA ($13\%$), and FinanceBench ($52\%$).
The primary metric is Exact Match (EM); secondary metrics include hallucination
rate and, for PAL configurations, first-attempt code compilation rate.

Table~\ref{tab:pipeline_2x2} reports the four-configuration sweep.  The full
VeNRA pipeline (Run 4: DualRetriever + PAL) achieves $49.1\%$ EM, a
$+9.9$\,pp absolute improvement over the baseline RAG configuration (Run 1:
$39.2\%$).  Hallucination rate simultaneously falls from $4.1\%$ to $1.2\%$,
reflecting the elimination of unconstrained free-form numerical generation in
favour of verifiable code execution.

\begin{table}[ht]
\centering
\caption{2$\times$2 pipeline evaluation.
Runs 1--2 use Gemini-3-flash CoT generation; Runs 3--4 use the VeNRA PAL
Code Agent.  Hallucination rate is computed as the fraction of incorrect answers
containing a numerically ungrounded value.  Compilation rate applies to PAL
runs only; it measures first-attempt Python execution success.}
\label{tab:pipeline_2x2}
\resizebox{\textwidth}{!}{%
\begin{tabular}{@{}clccccc@{}}
\toprule
\textbf{Run} &
\textbf{Configuration} &
\textbf{Retriever} &
\textbf{Generator} &
\textbf{EM $\uparrow$} &
\textbf{Halluc.\,\% $\downarrow$} &
\textbf{Compile\,\% $\uparrow$} \\ \midrule
1 & Baseline RAG
    & Vector  & Gemini CoT & $39.2\%$ & $4.1\%$ & --- \\
2 & Smart Ret., Dumb Math
    & VeNRA   & Gemini CoT & $37.4\%$ & $5.3\%$ & --- \\
3 & Dumb Ret., Smart Math
    & Vector  & VeNRA PAL  & $40.4\%$ & $1.8\%$ & $51.7\%$ \\
\midrule
\textbf{4} & \textbf{VeNRA Full (Ours)}
    & \textbf{VeNRA} & \textbf{VeNRA PAL}
    & $\mathbf{49.1\%}$ & $\mathbf{1.2\%}$ & $\mathbf{68.8\%}$ \\
\bottomrule
\end{tabular}}
\end{table}

\begin{figure}[htbp]
    \centering
    \includegraphics[width=0.82\textwidth]{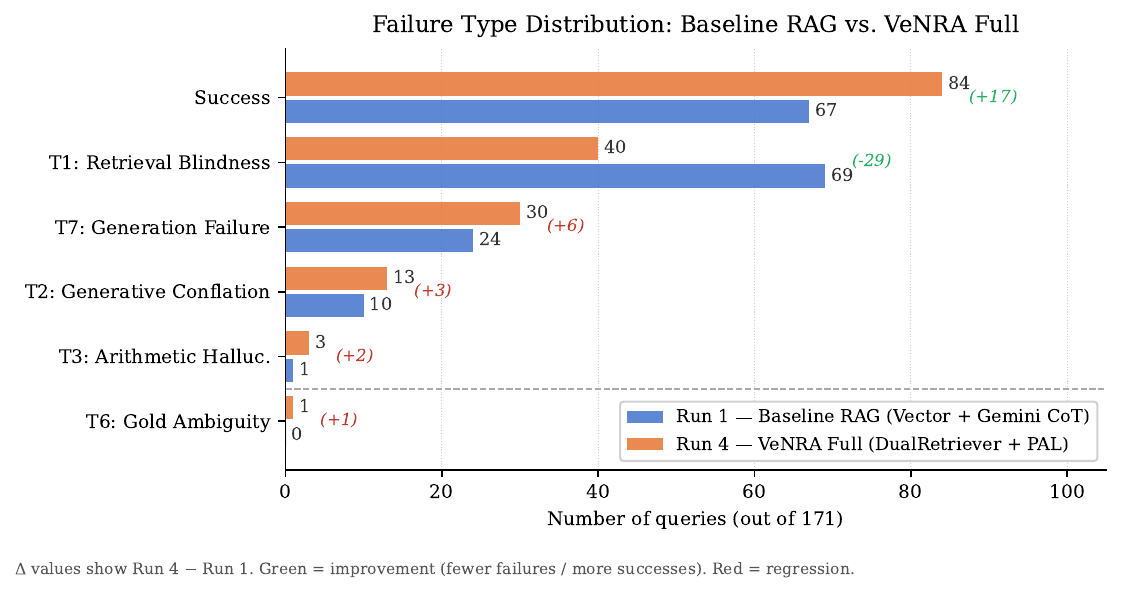}
    \caption{\textbf{Failure Type Transition: Run 1 (Baseline RAG) vs.\
    Run 4 (VeNRA Full)}.  Type-1 Retrieval Blindness falls by
    29 cases following Navigator rescue; Type-7 Generation Failures, which
    spiked under Run 2 (Smart Retrieval + CoT), return toward baseline levels
    once the PAL Code Agent replaces free-form generation.  The net gain of
    17 additional correct answers constitutes the $+9.9$\,pp EM improvement.}
    \label{fig:failure_taxonomy}
\end{figure}

Run 2 reveals a counterintuitive finding: replacing the baseline vector retriever
with the VeNRA Navigator \textit{degrades} EM from $39.2\%$ to $37.4\%$ when the
downstream generator remains a CoT model.
Failure analysis (Figure~\ref{fig:failure_taxonomy}) explains the mechanism.
The Navigator substantially reduces Type-1 Retrieval Blindness failures
($69 \to 35$ cases), recovering the correct evidence chunks for 34 previously
missed queries.  However, the enriched VeNRA context, averaging $1{,}236$ prompt tokens versus $680$ for the baseline ($1.8\times$ larger), due to the inclusion
of UFL-sourced numeric facts alongside retrieved chunks, simultaneously causes a
$+35$ spike in Type-7 Generation Failures.  Gemini-3-flash CoT, confronted with
dense, semi-structured financial tables interleaved with UFL fact rows, fails to
synthesise a well-formed answer for these additional cases, entirely offsetting
the retrieval gain.  This result has a direct architectural implication: the
VeNRA DualRetriever delivers correct evidence, but the generative bottleneck
must also be resolved, which is precisely the role of the PAL Code Agent.

The transition from Run 3 ($40.4\%$) to Run 4 ($49.1\%$) demonstrates that
VeNRA's two subsystems are positively complementary.  Improved retrieval
increases the compilation rate of the code agent from $51.7\%$ to $68.8\%$:
when the correct financial table is present in context, the code agent generates
syntactically valid programs that execute to the right answer at a significantly
higher rate.  The residual $31.2\%$ first-attempt compilation failure in Run 4
represents the remaining hard ceiling, driven primarily by queries requiring
cross-table joins or implicit unit conversions that exceed the agent's zero-shot
programming capacity.

The gains are not uniformly distributed across benchmarks.  On TAT-QA
($n{=}59$), EM is identical in Run 1 and Run 4 ($57.6\%$): the baseline vector
retriever already achieves adequate recall on TAT-QA's paragraph-bounded
questions (only 11 Type-1 failures in both runs), leaving no headroom for
Navigator-driven improvement.  The largest absolute gains occur on
FinanceBench ($24.4\% \to 40.0\%$, $+15.6$\,pp) and FinQA
($50.0\% \to 63.6\%$, $+13.6$\,pp), both of which require retrieving sparse
numerical evidence from long-form financial filings, precisely the regime where
the Navigator's first-pass miss recovery ($88.6\%$ rescue rate) provides the
greatest lift.

Figure~\ref{fig:failure_taxonomy} presents the per-type failure counts for Run 1
(Baseline) and Run 4 (VeNRA Full).  The dominant improvement is the
$29$-case reduction in Type-1 Retrieval Blindness ($69 \to 40$), attributable
to the Navigator's ability to execute a targeted re-retrieval pass when the
initial vector search fails.  Type-2 Generative Conflation modestly increases
($10 \to 13$), consistent with VeNRA Full surfacing previously-inaccessible
queries that are difficult for even a code-based generator: these cases expose
multi-entity aggregation requirements where the correct evidence is retrieved but
the program conflates semantically similar line items (e.g., \textit{operating
income} vs.\ \textit{income from operations}).  The survivor analysis identifies
$13$ queries ($7.6\%$) that are solved \textit{exclusively} by VeNRA Full, cases
where Run 1 fails on all retries, and $64$ ($37.4\%$) that remain unsolved by
all configurations, representing the intrinsic difficulty floor of the benchmark
under zero-shot conditions.

\subsection{Experiment 2: Ingestion Purity and Double-Lock Filtering}
\label{subsec:exp2_ingestion}

A prerequisite for grounded numerical reasoning is a knowledge base whose
entries are verified against their source documents.  Unverified or mis-aligned
facts injected into the retrieval context produce a failure mode we term
\textit{UFL Bleed}: the model anchors on a plausible but incorrect numerical
claim, generating a fluent but hallucinated answer that is invisible to
surface-level quality checks.

VeNRA addresses this through a two-stage \textit{Double-Lock} filtering
pipeline applied during ingestion.  \textit{Lock 1} (Proposal) uses a
zero-shot extraction model to propose candidate text facts from each chunk,
yielding a set of fact candidates with an associated alignment hypothesis.
\textit{Lock 2} (Verification) re-reads each candidate against its source
chunk, independently confirming the numerical value, entity binding, and
temporal scope.  Only facts passing both stages are committed to the
UFL.

Table~\ref{tab:ingestion_purity} summarises the pipeline over the full
$899$-chunk ingestion run.  Lock 1 proposed $2{,}541$ text fact candidates;
Lock 2 rejected $754$ of these, yielding a Double-Lock rejection rate of
$29.7\%$.  The $1{,}787$ accepted text facts, combined with $1{,}278$ structured
numeric facts extracted via a separate deterministic parser, constitute the
final UFL of $3{,}065$ rows.

\begin{table}[ht]
\centering
\caption{Double-Lock ingestion statistics over 899 document chunks.
Text facts and numeric facts follow separate extraction paths; the Double-Lock
filter applies only to the text fact track.  Alignment tier and confidence
score are assigned at Lock 2 verification.}
\label{tab:ingestion_purity}
\begin{tabular}{@{}lrr@{}}
\toprule
\textbf{Metric} & \textbf{Count / Value} & \textbf{Share} \\ \midrule
Chunks processed                  & $899$   & --- \\
Text facts proposed (Lock 1)      & $2{,}541$ & --- \\
Text facts rejected (Lock 2)      & $754$   & $29.7\%$ \\
Text facts accepted               & $1{,}787$ & $70.3\%$ \\
Numeric facts (deterministic)     & $1{,}278$ & --- \\
\midrule
\textbf{UFL total rows}           & $\mathbf{3{,}065}$ & --- \\
\midrule
\multicolumn{3}{@{}l}{\textit{Alignment tier (final UFL)}} \\
\quad EXACT match                 & $2{,}282$ & $74.5\%$ \\
\quad PARTIAL match               & $750$   & $24.5\%$ \\
\quad FUZZY match                 & $33$    & $1.1\%$  \\
\midrule
\multicolumn{3}{@{}l}{\textit{Confidence score (Lock 2)}} \\
\quad Mean                        & $0.793$ & --- \\
\quad Median                      & $0.700$ & --- \\
\quad 75th percentile             & $0.950$ & --- \\
\quad Min / Max                   & $0.610$ / $0.950$ & --- \\
\bottomrule
\end{tabular}
\end{table}

The EXACT tier ($74.5\%$ of rows) indicates that the extracted numerical value,
entity label, and period specification are verbatim-matched to a continuous span
in the source chunk.  PARTIAL alignment ($24.5\%$) covers facts where the value
is correctly extracted but the entity scope or temporal boundary required
inference across multiple sentences, these are the facts most vulnerable to
UFL Bleed in the absence of Lock 2 verification.  FUZZY alignment ($1.1\%$)
arises when the source chunk uses a reformulation (e.g.\ a percentage expressed
as a decimal) that Lock 1 normalises; these facts carry the lowest mean
confidence and are used only when no EXACT or PARTIAL alternative exists for the
queried metric.

The bimodal confidence distribution (median $0.700$, 75th percentile $0.950$)
reflects a natural separation between facts that Lock 2 confidently verifies
($\text{conf}=0.95$, assigned to unambiguous numeric spans) and facts requiring
cross-sentence inference ($\text{conf}\in[0.61, 0.70]$).  No fact with
$\text{conf}<0.61$ is admitted to the UFL, providing a hard floor on factual
precision at the cost of recall on ambiguously stated financial metrics.

Across the full evaluation, only one instance of
Type-0 UFL Bleed was observed (Run 2, Table~\ref{tab:pipeline_2x2}), a case
where a PARTIAL, alignment fact for a structurally similar company was retrieved
alongside the correct evidence, and the CoT generator anchored on it.
This single occurrence ($0.6\%$ of queries) confirms that the Double-Lock
protocol achieves near-complete contamination containment under the evaluation
conditions, at the cost of the $29.7\%$ proposed-fact rejection rate documented
above.

\subsection{Experiment 3: Sentinel Forensic Auditing}
\label{subsec:exp3_sentinel}

To validate the VeNRA Sentinel and our Adversarial Simulation strategy, we evaluate the trained model against its zero-shot base and a suite of frontier LLMs spanning
proprietary and open-weight systems.  The primary objective is to measure discriminative
capacity on \textit{Hard Negatives} without inducing \textit{Paranoid Sycophancy}, the
pathological tendency to maximise detection recall by indiscriminately labelling all inputs
as \texttt{UNFOUNDED}.

We introduce the \textit{Composite Score} ($\mathcal{M}$) as out \textbf{Evaluation Metrics}. It is a multiplicative metric that
jointly penalises any failure in the four orthogonal auditing capabilities:
\begin{equation}
    \mathcal{M} = \sqrt{\,
        \widetilde{\mathrm{FR}}_{\mathrm{global}} \;\times\;
        \widetilde{\mathrm{Recall}}_{\mathrm{nat}} \;\times\;
        \widetilde{\mathrm{TPR}}_{\mathrm{clean}} \;\times\;
        \widetilde{\mathrm{Acc}}_{\mathrm{axiom}}
    }\,,
    \label{eq:composite}
\end{equation}
where each term is a Jeffreys-smoothed posterior mean and the root normalises the
scale.  $\mathrm{FR}_{\mathrm{global}}$ is the \textit{Paired Flip Rate}: both the
original (\texttt{SUPPORTED}) and its programmatically sabotaged child (\texttt{UNFOUNDED})
must be classified correctly for a pair to count.
$\mathrm{Recall}_{\mathrm{nat}}$ is the detection rate on organically occurring (non-synthetic)
hallucinations.
$\mathrm{TPR}_{\mathrm{clean}}$ is the True Positive Rate on valid, un-sabotaged documents
(resistance to false alarms).
$\mathrm{Acc}_{\mathrm{axiom}}$ measures retention of parametric world knowledge on the
\texttt{GENERAL} class.  Because $\mathcal{M}$ is multiplicative, a model that collapses any
single pillar, e.g.\ by blindly predicting \texttt{UNFOUNDED}, is driven to near-zero
regardless of its performance on the remaining three.

\begin{figure}[htbp]
    \centering
    \begin{subfigure}[b]{0.55\textwidth}
        \includegraphics[width=\textwidth]{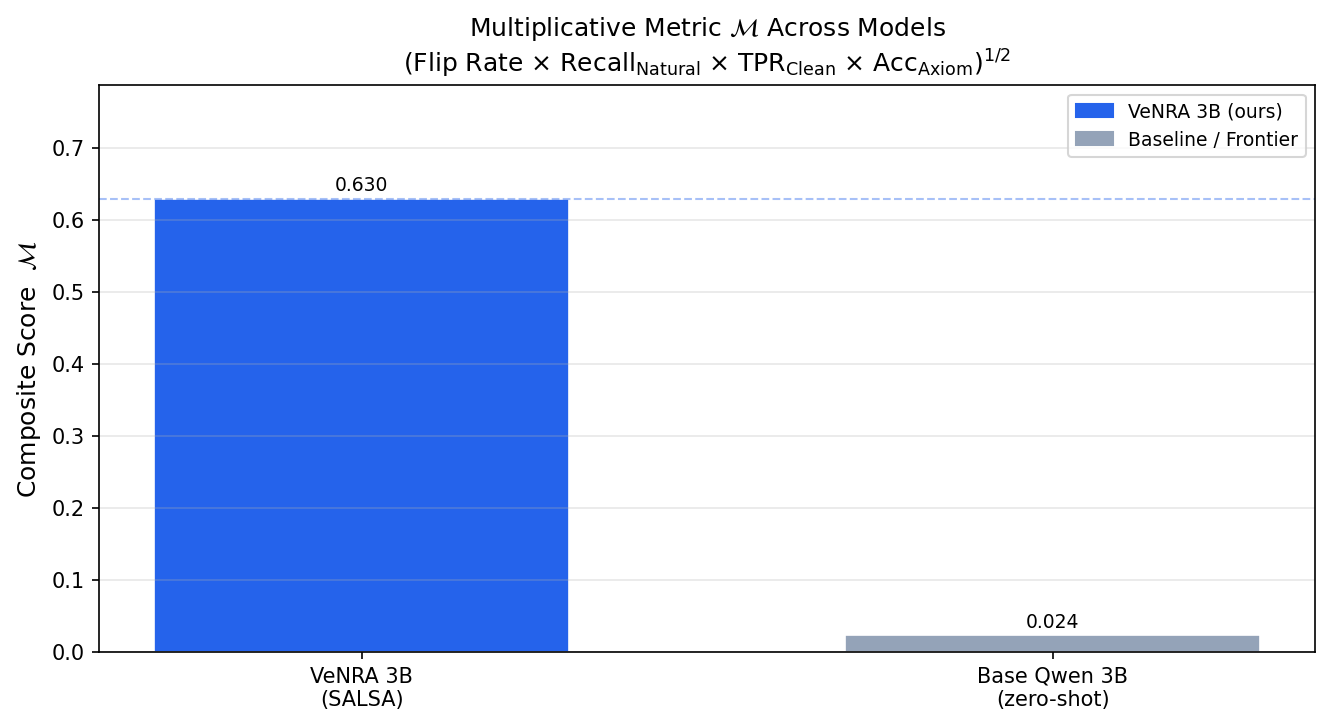}
        \caption{Composite Score $\mathcal{M}$ comparison.}
        \label{fig:composite_bar}
    \end{subfigure}
    \hfill
    \begin{subfigure}[b]{0.43\textwidth}
        \includegraphics[width=\textwidth]{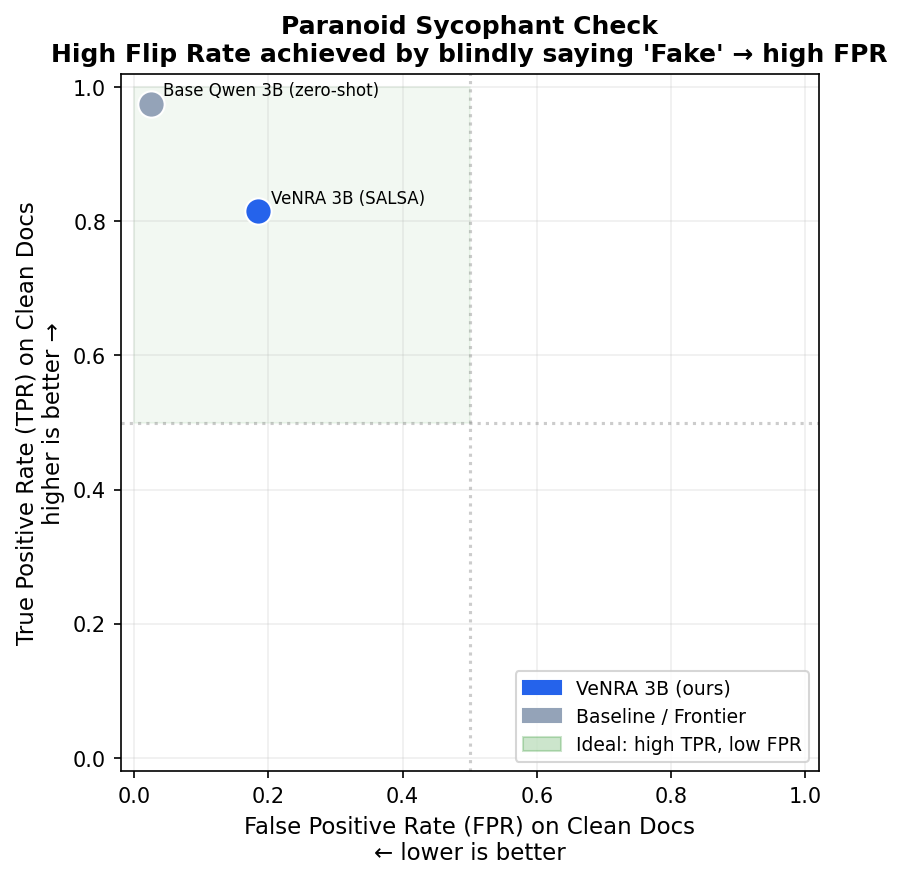}
        \caption{FPR–TPR operating points.}
        \label{fig:fpr_tpr}
    \end{subfigure}
    \caption{\textbf{Full-distribution auditing performance.} Left: $\mathcal{M}$ scores
    on the test set, demonstrating a $26.5\times$ improvement.  Right: the
    FPR–TPR plane reveals the sycophancy trap, high TPR is trivially achievable by
    suppressing all detections, rendering the model operationally useless. The shaded green region marks the
        high-utility quadrant (low false-alarm rate, high valid-claim acceptance).
        The base model achieves high TPR only by never firing.}
\end{figure}

Table~\ref{tab:main_auditing} reports the complete evaluation on the test set,  where all four pillars of $\mathcal{M}$ are populated.
The zero-shot base model exhibits catastrophic confirmation bias: it achieves a
near-perfect $\mathrm{TPR}_{\mathrm{clean}}$ of $97.5\%$ by indiscriminately classifying
almost every input as \texttt{SUPPORTED}, causing its Flip Rate to collapse to $5.3\%$ and
its natural hallucination recall to $2.1\%$.  The resulting composite score is
$\mathcal{M}=0.024$.  Our trained Sentinel overcomes this prior, reaching a
$93.0\%$ Global Flip Rate and $83.6\%$ natural recall, while sustaining a $81.5\%$
$\mathrm{TPR}_{\mathrm{clean}}$, demonstrating that the model has learned a calibrated
detection threshold rather than a degenerate majority-class prior.
The $26.5\times$ improvement in $\mathcal{M}$ ($0.630$ vs.\ $0.024$) is further
visualised in Figure~\ref{fig:composite_bar}.

Notably, the Sentinel exhibits \textit{length-invariance}: Flip Rate on long-context sequences
($\geq512$ tokens, $n=79$ pairs) is $93.7\%$, marginally higher than on short contexts
($92.6\%$, $n=149$ pairs), refuting the hypothesis that evidence grounding degrades under
extended contexts.  Figure~\ref{fig:fpr_tpr} plots the FPR–TPR operating points for both
models, visualising the sycophancy axis: the base model occupies the top-right corner of the
\textit{paranoid sycophant} region (high TPR achieved only via universal acceptance), while
the Sentinel sits in the high-utility top-left quadrant.

\begin{table}[ht]
\centering
\caption{Full test-set evaluation ($n{=}812$). $\mathcal{M}$ is computable only here, where
all four metric pillars are populated. Axiom accuracy is indicative only ($n{=}10$ held-out
axioms). \textbf{Bold} denotes best per column; $\uparrow$/$\downarrow$ indicate preferred
direction.}
\label{tab:main_auditing}
\resizebox{\textwidth}{!}{%
\begin{tabular}{@{}lcccccccc@{}}
\toprule
\textbf{Model} &
\textbf{Valid\%} &
\textbf{FR$_{S}$ $\uparrow$} &
\textbf{FR$_{L}$ $\uparrow$} &
\textbf{FR$_{\mathrm{global}}$ $\uparrow$} &
\textbf{Recall$_{\mathrm{nat}}$ $\uparrow$} &
\textbf{TPR$_{\mathrm{clean}}$ $\uparrow$} &
\textbf{FPR$_{\mathrm{clean}}$ $\downarrow$} &
\textbf{$\mathcal{M}$ $\uparrow$} \\ \midrule
Base Qwen 3B (Zero-Shot)
    & $100.0\%$ & $2.7\%$  & $10.1\%$ & $5.3\%$  & $2.1\%$
    & $\mathbf{97.5\%}$ & $\mathbf{2.5\%}$ & $0.024$ \\
\textbf{VeNRA 3B (Ours)}
    & $\mathbf{100.0\%}$ & $\mathbf{92.6\%}$ & $\mathbf{93.7\%}$ & $\mathbf{93.0\%}$
    & $\mathbf{83.6\%}$ & $81.5\%$ & $18.5\%$
    & $\mathbf{0.630}$ \\ \bottomrule
\end{tabular}}
\end{table}

To contextualise the Sentinel, we conducted a stratified evaluation  from all four sabotage typologies.  Because this subsample contains only adversarial pairs and no natural, failure or axiom instances by design, $\mathcal{M}$ reduces to the Flip Rate alone.  Table~\ref{tab:frontier_sabotage} presents the
results.

A critical methodological distinction must be noted: models with mandatory reasoning preambles
(Gemini-3-Flash-Preview, GPT-OSS-120B) and smaller models with weak instruction following (Qwen3-32B) were allocated up to $100$ generation
tokens before the verdict token, to prevent format-induced parse failures.  All remaining
models, including VeNRA, were constrained to a strict \textbf{single token budget}. The validity rate column quantifies an underreported operational risk of deploying frontier
models as binary classifiers.  Gemini-3-Flash-Preview failed to emit a parseable verdict
on $37.0\%$ of queries despite its relaxed token budget ($63.0\%$ valid), a production
failure rate that would be unacceptable in any real-time auditing pipeline.
GPT-OSS-120B achieved only $79.4\%$ validity for the same reason.

\begin{table}[ht]
\centering
\caption{Adversarial subsample evaluation ($n{=}92$ rows, 46 pairs; ${\sim}13$ per Logic/Numeric,
${\sim}12$ Irrelevancy, ${\sim}8$ Semantic Drift). Models marked $\dagger$ were allocated 100 generation tokens budget to response; all others (except CoT) operate under a strict
1-token constraint. Global FR and per-type rates are computed only over valid responses.
\textbf{Bold}: best per column.}
\label{tab:frontier_sabotage}
\resizebox{\textwidth}{!}{%
\begin{tabular}{@{}llccccccc@{}}
\toprule
\textbf{Model} &
\textbf{Tokens} &
\textbf{Valid\%} &
\textbf{Global FR $\uparrow$} &
\textbf{Logic Code Lie} &
\textbf{Numeric Neighbor} &
\textbf{Irrelevancy RAG} &
\textbf{Semantic Drift} \\ \midrule
Base Qwen 3B (Zero-Shot)         & 1      & $100.0\%$ & $8.7\%$  & $6.1\%$  & $0.0\%$           & $8.0\%$           & $0.0\%$  \\
Base Qwen 3B ( + CoT) & 150 & $97.8\%$  & $15.2\%$ & $15.4\%$ & $15.4\%$          & $25.0\%$          & $0.0\%$  \\
Gemini-3-Flash-Preview$^\dagger$ & 100    & $63.0\%$  & $39.1\%$ & $15.4\%$ & $46.2\%$          & $75.0\%$          & $12.5\%$ \\
GPT-OSS-120B $^\dagger$   & 100    & $79.4\%$  & $56.5\%$ & $61.5\%$ & $92.3\%$          & $41.7\%$          & $12.5\%$ \\
Kimi K2.5           & 1      & $93.96\%$ & $69.6\%$ & $69.2\%$ & $92.3\%$          & $91.7\%$          & $0.0\%$  \\
Llama 3.3 70B            & 1      & $100.0\%$ & $71.7\%$ & $76.9\%$ & $84.6\%$          & $\mathbf{100.0\%}$& $0.0\%$  \\
Qwen3-32B $^\dagger$      & 100    & $100.0\%$ & $73.9\%$ & $84.6\%$ & $84.6\%$          & $91.7\%$          & $12.5\%$ \\
Gemini-2.5-Flash                & 1      & $100.0\%$ & $80.4\%$ & $84.6\%$ & $\mathbf{92.3\%}$ & $83.3\%$          & $50.0\%$ \\ \midrule
\textbf{VeNRA 3B (Ours)}  & \textbf{1} & $\mathbf{100.0\%}$ & $\mathbf{91.3\%}$ & $\mathbf{92.3\%}$ & $\mathbf{92.3\%}$ & $\mathbf{100.0\%}$ & $\mathbf{75.0\%}$ \\ \bottomrule
\end{tabular}}
\end{table}

\begin{figure}[htbp]
    \centering
    \includegraphics[width=0.72\textwidth]{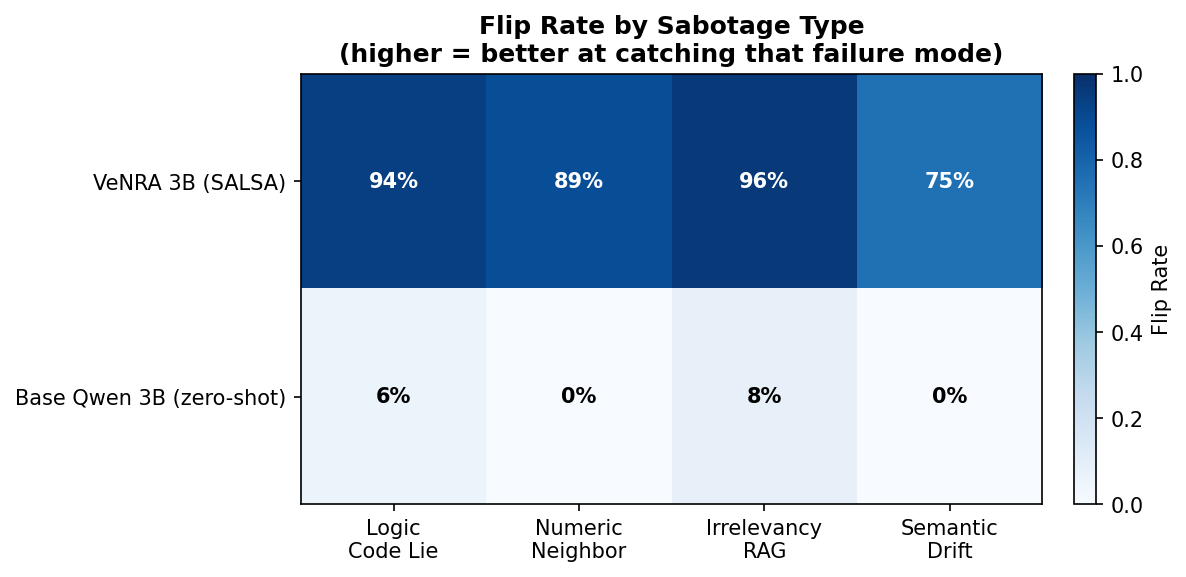}
    \caption{\textbf{Detection Rate Across Sabotage Typologies.}
    The Sentinel achieves ${\geq}88\%$ detection across all four forensic failure modes.
    The base model's near-zero rates across all typologies confirm that the discriminative
    capability is entirely attributable to fine-tunning, not to the base model's
    pre-trained knowledge.}
    \label{fig:sabotage_heatmap}
\end{figure}

Three patterns emerge from Table~\ref{tab:frontier_sabotage}.  First, the \texttt{base\_qwen\_cot}
row demonstrates that test-time Chain-of-Thought reasoning provides only marginal uplift over
zero-shot classification ($8.7\% \to 15.2\%$ Global FR), confirming that the discriminative
capability originates from fine-tuning rather than from reasoning token budget.
Second, the \textit{Semantic Drift} column exposes a universal blind spot: five of seven
frontier models score $0.0\%$ on this typology, including Llama 3.3 70B and Kimi K2.5
despite strong performance elsewhere.  Semantic Drift sabotages replace a semantically
anchored entity (e.g.\ \textit{operating income}) with a plausible near-synonym (e.g.\
\textit{operating profit}), a substitution that passes surface plausibility checks but
violates the formal financial taxonomy.  Only VeNRA ($75.0\%$) and Gemini-2.5-Flash
($50.0\%$) exhibit meaningful detection, suggesting this typology requires explicit
training signal rather than generative reasoning.
Third, the Logic Code Lie column—where a Python execution trace is syntactically
flawless but input variables are swapped, shows the largest discrimination between trained
and untrained models.  Gemini-3-Flash, despite having a $39.1\%$ Global FR, achieves only
$15.4\%$ on Logic Code Lies (lower than the zero-shot 3B base), indicating that its
chain-of-thought process is actively misled by the surface coherence of the fabricated trace.
Figure~\ref{fig:sabotage_heatmap} visualises the per-typology detection rates for VeNRA
and the zero-shot base on the full test set.

\subsection{Experiment 4: Systems Efficacy and Training Dynamics}
\label{subsec:exp4_systems}

Real-time financial auditing imposes strict latency constraints.  This experiment
quantifies (i) the inference speedup conferred by the architecture versus standard
Chain-of-Thought, and (ii) the training dynamics that make calibration
achievable under extreme differential loss penalisation.

\begin{figure}[htbp]
    \centering
    \begin{subfigure}[b]{0.47\textwidth}
        \includegraphics[width=\textwidth]{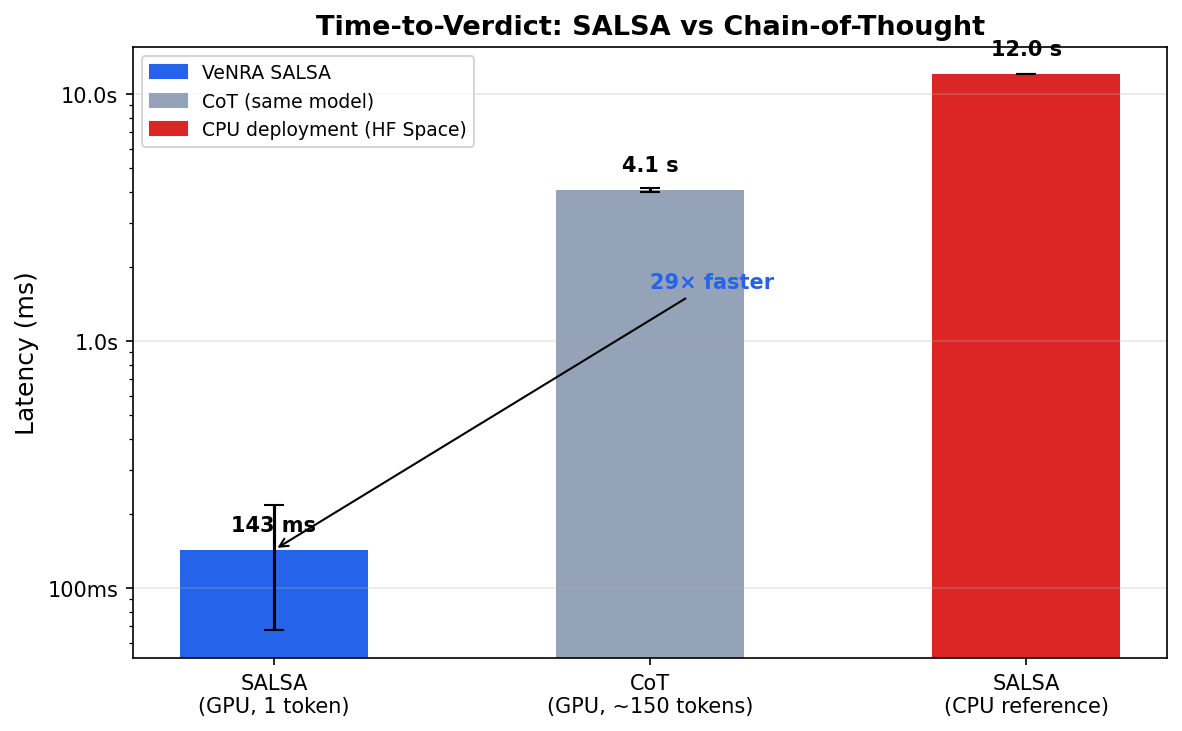}
        \caption{Median Time-to-Verdict: VeNRA vs.\ CoT on GPU, and a CPU reference
        deployment scenario (not a direct comparison).}
        \label{fig:latency_bar}
    \end{subfigure}
    \hfill
    \begin{subfigure}[b]{0.47\textwidth}
        \includegraphics[width=\textwidth]{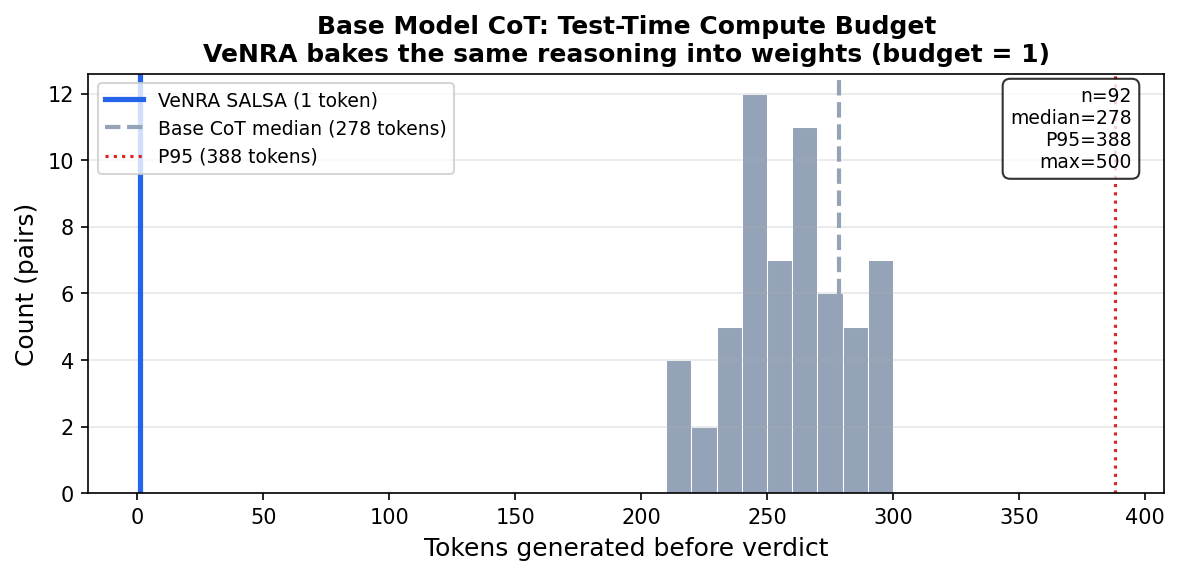}
        \caption{Distribution of tokens generated before verdict across our frontier
        evaluation suite (median $278.5$, P95 $388.2$). VeNRA collapses this to a
        fixed budget of $\mathbf{1}$ token.}
        \label{fig:cot_budget}
    \end{subfigure}
    \caption{\textbf{Latency Efficiency.}  Standard CoT requires hundreds of
    reasoning tokens before committing to a verdict; VeNRA encodes the decision into the
    first generated token, achieving a $28.7\times$ reduction in median latency.}
    \label{fig:latency_combined}
\end{figure}

We measured median Time-to-Verdict on $n{=}100$ prompts (50 short, 50 long) under
identical hardware conditions (NVIDIA A40).  Standard Chain-of-Thought generation was
capped at 150 tokens to ensure a finite measurement; however, our frontier evaluation
shows that uncapped CoT generation reaches a median of $278.5$ tokens before emitting
a verdict (mean $289.0$, P95 $388.2$), making the 150-token benchmark a conservative
lower bound on real-world latency. Results are reported in Table~\ref{tab:latency} and visualised in
Figures~\ref{fig:latency_bar}--\ref{fig:cot_budget}.
VeNRA achieves a median verdict latency of $\mathbf{142.6}$\,ms (P95: $217.8$\,ms),
compared to $4{,}092.5$\,ms for capped CoT (P95: $4{,}178.6$\,ms), a
$\mathbf{28.7\times}$ acceleration.  Extrapolating to uncapped CoT at the empirically
observed median of $278.5$ tokens, the proportional estimate is ${\sim}7{,}594$\,ms,
implying a real-world speedup exceeding $53\times$.
In CPU-bound edge deployments (reference: ${\sim}12{,}000$\,ms per verdict under CoT,
derived from our $\sim$10\,s/token CPU throughput estimate), the SALSA paradigm
reduces verification latency from an operationally unviable ${\sim}25$\,minutes to
${\sim}10$\,seconds, enabling real-time rather than batch-mode auditing.

\begin{table}[ht]
\centering
\caption{Time-to-Verdict latency benchmark ($n{=}100$ prompts, NVIDIA A40).
CoT was capped at 150 generation tokens; see text for uncapped extrapolation.}
\label{tab:latency}
\begin{tabular}{@{}lcccc@{}}
\toprule
\textbf{Condition} & \textbf{Median (ms)} & \textbf{Mean (ms)} & \textbf{P95 (ms)} & \textbf{Speedup} \\
\midrule
CoT (${\leq}150$ tok, capped)  & $4{,}092.5$ & $4{,}017.0$ & $4{,}178.6$ & $1.0\times$ (baseline) \\
VeNRA (1 tok)                  & $\mathbf{142.6}$ & $\mathbf{150.8}$ & $\mathbf{217.8}$ & $\mathbf{28.7\times}$ \\
\bottomrule
\end{tabular}
\end{table}

\subsection{Solving Gradient Starvation and Mode Collapse}
\label{subsec:exp5_systems}
Training a model to emit a calibrated verdict as its \textit{first} autoregressive token
introduces compounding instabilities.  Table~\ref{tab:progressive_ablation} details our
progressive ablation over the Micro-Chunking loss architecture; each row isolates the
marginal contribution of one design decision, evaluated at a fixed intermediate checkpoint
to ensure comparability\footnote{The ablation rows reflect evaluations at a fixed early
checkpoint. The fully converged model (Table~\ref{tab:main_auditing}) achieves
$\mathcal{M}{=}0.630$ after complete training, compared to $0.201$ at the ablation
checkpoint; the $3.1\times$ gap reflects continued improvement under cosine annealing.}. The ablation reveals three distinct failure modes, each resolved by a targeted
architectural modification.

\begin{table}[ht]
\centering
\caption{Progressive ablation of the Micro-Chunking loss architecture, evaluated at a
fixed intermediate checkpoint. $\mathcal{M}$, Flip Rate, per-class accuracies, TPR/FPR
on clean data, and ECE are reported. Recall$_{\mathrm{nat}}$ is included where
estimable at this checkpoint. Rows without $\mathcal{M}$ indicate configurations where
training was terminated before reaching a measurable composite score (gradient explosion
or class collapse).}
\label{tab:progressive_ablation}
\resizebox{\textwidth}{!}{%
\begin{tabular}{@{}lcccccccccc@{}}
\toprule
\textbf{Configuration} &
\textbf{$\mathcal{M}$ $\uparrow$} &
\textbf{FR $\uparrow$} &
\textbf{Found $\uparrow$} &
\textbf{Fake $\uparrow$} &
\textbf{Axiom $\uparrow$} &
\textbf{Recall$_{\mathrm{nat}}$ $\uparrow$} &
\textbf{TPR $\uparrow$} &
\textbf{FPR $\downarrow$} &
\textbf{ECE $\downarrow$} \\ \midrule
Baseline (Standard CE)
    & 0 & $0.115$ & $0.894$ & $0.192$ & ---   & ---   & ---   & ---   & ---   \\
$+$ Weighted ($50{\times}$), No Clamp
    & 0.002 & $0.295$ & $0.657$ & $0.581$ & ---   & ---   & ---   & ---   & ---   \\
$+$ Clamped, Uniform ($50/50/50$)
    & $0.007$ & $0.229$ & $0.739$ & $0.299$ & $0.286$ & $0.150$ & $0.700$ & $0.063$ & $0.439$ \\
$+$ Differential ($50/50/10$)
    & $0.032$ & $0.325$ & $0.637$ & $0.452$ & $0.571$ & $0.275$ & $0.625$ & $0.138$ & $0.415$ \\
\midrule
$+$ Prompt Tuning \textbf{(Final)}
    & $\mathbf{0.201}$ & $\mathbf{0.363}$ & $\mathbf{0.777}$ & $\mathbf{0.433}$ & $\mathbf{0.643}$ & $0.263$ & $\mathbf{0.725}$ & $\mathbf{0.125}$ & $\mathbf{0.375}$ \\ \bottomrule
\end{tabular}}
\end{table}

\textit{(i) Gradient Starvation.}  Standard unweighted Cross-Entropy (Baseline) fails
to converge on the decision boundary.  Because the single verdict token's gradient
contribution is diluted by the subsequent ${\sim}278$ reasoning tokens in the loss
window, the model optimises for fluent rationale generation while ignoring the
discriminative token, exhibiting near-sycophantic behaviour (Flip Rate $11.5\%$,
Fake Accuracy $19.2\%$).

\begin{figure}[htbp]
    \centering
    \includegraphics[width=0.7\textwidth]{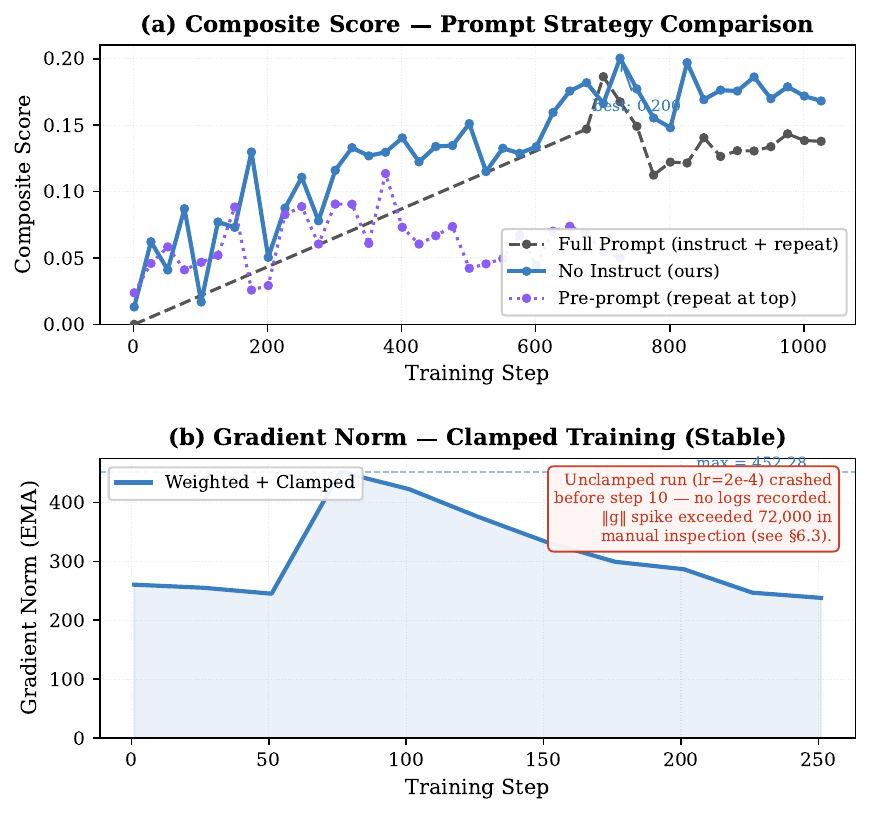}
    \caption{\textbf{Training Dynamics under Micro-Chunking Loss.}
    \textit{(a)} Composite Score trajectory across prompt strategies; peak $\mathcal{M}{=}0.201$
    at this checkpoint, rising to $0.630$ at full convergence.
    \textit{(b)} Gradient norm (EMA) under the final clamped configuration; peak $452.3$
    confirms stable training.  The unclamped crashed
    with gradient norms exploding.}
    \label{fig:training_dynamics}
\end{figure}

\textit{(ii) Gradient Explosion.}  Applying a naive $50{\times}$ penalty to the verdict
token without dynamic loss clamping induced catastrophic instability.  Manual inspection
of the forward pass immediately prior to the run crashing before step 10 revealed gradient
norms exploded.  Micro-Chunking with dynamic clamping
($\ell_{\max} = 250$) resolves this by bounding per-token loss contributions, yielding
stable gradient norms throughout (maximum observed EMA: $452.3$, as shown in
Figure~\ref{fig:training_dynamics}).

\textit{(iii) Minority-Class Probability Absorption (``Garbage Bin'' Effect).}
With loss clamping applied but \textit{uniform} class weighting ($50/50/50$), the model
discovers a low-loss refuge in the minority \texttt{GENERAL} class, collapsing
\texttt{Fake} accuracy to $29.9\%$ while \texttt{GENERAL} accuracy rises to $28.6\%$—
evidence that probability mass migrates to the axiomatic class to avoid the steep
penalisation on the core binary decision boundary.  Differential weighting ($50/50/10$)
eliminates this bleed: reducing the \texttt{GENERAL} penalty forces the model to
re-engage with the \texttt{Found}/\texttt{Fake} boundary, recovering Fake accuracy to
$45.2\%$ and Axiom accuracy to $57.1\%$ simultaneously.

\begin{figure}[htbp]
    \centering
    \includegraphics[width=0.85\textwidth]{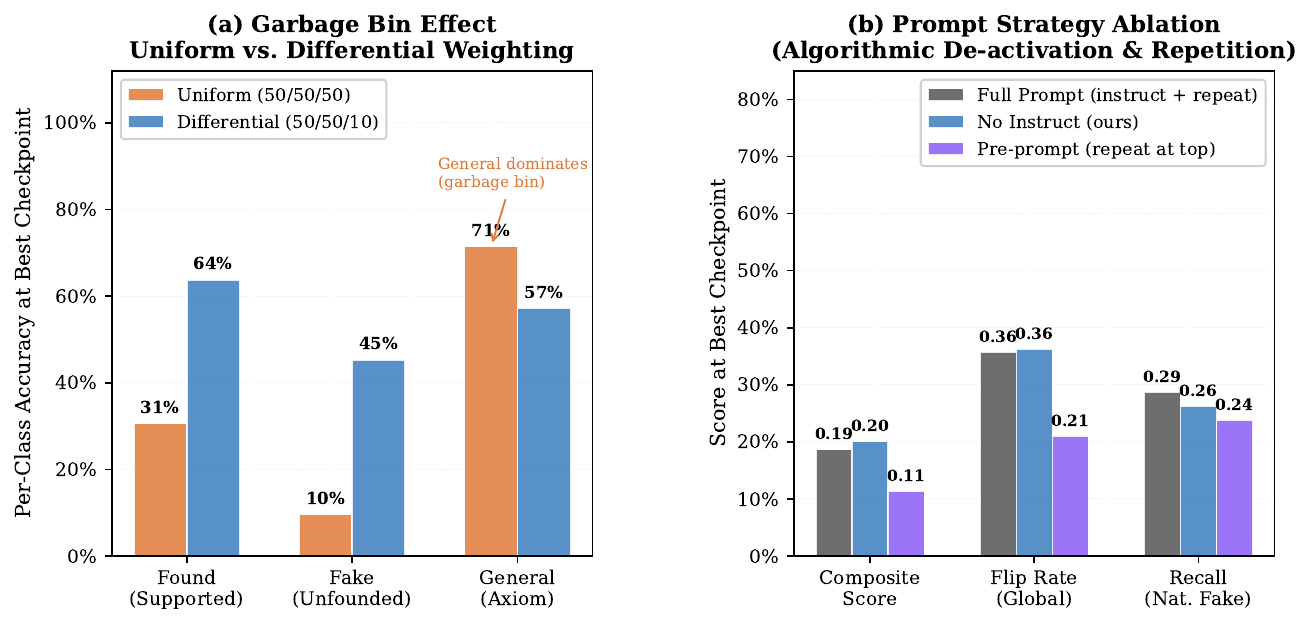}
    \caption{\textbf{Training Pathology Visualisation.}
    \textit{(a) Garbage Bin Effect:} Per-class accuracy under uniform vs.\ differential
    weighting.  Uniform weighting causes probability mass to collapse into the minority
    \texttt{GENERAL} class; differential weighting ($50/50/10$) recovers the core binary
    boundary.
    \textit{(b) Prompt Strategy Ablation:} Removing instruction text (``No Instruct'')
    marginally improves or matches full-prompt performance (Composite $0.200$ vs.\ $0.190$), compare to full prompt due,
    while front-loading instruction degrades it significantly ($0.110$).}
    \label{fig:class_ablation}
\end{figure}

A complementary ablation over three prompt configurations, \textit{Full Prompt} (direct instruction with repetition), \textit{No Instruct} (repetition without direct instruction),
and \textit{Pre-prompt} (direct instruction with no repetition), reveals that the
\textit{No Instruct} variant achieves the highest composite score ($0.200$) at
comparable Flip Rate ($36\%$) to the full prompt ($0.190$, $36\%$).
Pre-prompt repetition degrades performance sharply ($0.110$, $21\%$), likely by
dilution of the trace, candidate vs context.
Collectively, these results indicate that the task capability is encoded primarily in
the training distribution rather than the inference-time prompt, which is consistent
with the SALSA hypothesis that decision knowledge for classification should reside in model weights rather than in generation-time reasoning chains.

\section{Discussion, Limitations, and Conclusion}
\label{sec:Discussion}

\subsection{Discussion: The Neuro-Symbolic Efficiency Frontier}
The prevailing trajectory in Large Language Model research assumes that hallucination mitigation is a function of parametric scaling. The VeNRA framework demonstrates an alternative, highly efficient paradigm: \textit{Neuro-Symbolic Cognitive Offloading}. By acknowledging that LLMs are fundamentally probabilistic sequence generators, VeNRA surgically removes their responsibility to perform arithmetic, bounding them instead within strictly typed schemas (the UFL) and deterministic execution environments (Python). 

This architectural shift yields a dual benefit. First, it achieves ``Zero-Hallucination'' arithmetic by construction, provided the initial extraction is grounded. Second, the development of \textbf{VeNRA-Data} proves that the bottleneck in creating reliable ``LLM-as-a-Judge'' models is not model size, but data ecology. By shifting evaluation benchmarks from linguistically obvious semantic noise to structurally precise \textit{Adversarial Simulation} (e.g., Logic Code Lies), we empower locally deployable $3$B parameter SLMs to achieve forensic verification capabilities previously restricted to frontier models, effectively democratizing high-stakes financial auditing.

\subsection{Limitations}
Despite its structural rigor, the VeNRA architecture possesses inherent limitations grounded in its reliance on deterministic extraction:
\begin{enumerate}
    \item \textbf{Recursive Graph Blindness:} The UFL flattens financial reports into a relational table. While this enables rapid SQL-like filtering, it fails to capture recursive second-order relationships (e.g., ``Supplier of a Supplier'') inherent in complex supply chains. Currently, the Naive Graph Expansion relies on single-step filtering, potentially missing multi-hop risks.
    \item \textbf{Referential Nuance Loss:} Financial footnotes frequently employ referential pointers (e.g., ``See Note 12''). While our \textit{Trailing-Buffer Chunking} captures immediate context, a footnote extracted into the \texttt{text\_nuance} column loses its hypertextual integrity. If the referenced note resides in a disjointed chunk not retrieved by the Navigator, the qualitative context is severed, potentially leading to misinterpretation of conditional covenants.
    \item \textbf{The Semantic Schema Bottleneck:} The Lexical Pre-Filter effectively prevents vector conflation for standard GAAP metrics. However, highly idiosyncratic corporate taxonomies (e.g., non-GAAP ``Adjusted EBITDAR excluding Stock-Based Comp'') may evade the Navigator's synonym mapping. If the Lexical Gate is too aggressive, it risks filtering out valid but non-standard metric terminologies.
\end{enumerate}

\subsection{Conclusion}
\label{sec:conclusion}

VeNRA establishes a novel blueprint for grounded RAG
in financial reasoning by addressing the two root causes of RAG failure,
retrieval conflation and arithmetic hallucination, at the architectural level
rather than through post-hoc filtering. The \textbf{UFL} replaces probabilistic text retrieval with
deterministic typed variables, and the \textbf{Double-Lock Grounding} protocol
acts as a strict extraction firewall before they can contaminate the retrieval context.
The PAL Code Agent then eliminates free-form numerical generation entirely,
reducing hallucination rates to $1.2\%$ on a cross-benchmark
evaluation spanning FinanceBench, FinQA, and TAT-QA.

The \textbf{VeNRA Sentinel} addresses the residual error surface through
forensic trace auditing.
Trained exclusively on programmatically constructed \textit{Ecological Errors}
rather than naturally occurring hallucination corpora, the 3B-parameter Sentinel
achieves $>91\%$ adversarial Flip Rate on the controlled evaluation—
outperforming all SoTA model
while operating within a ${\lesssim}150$\,ms inference budget at the $28{\times}$
speedup delivered by single-token classification.

Three contributions emerge as independently transferable to other domains:
the Micro-Chunking loss algorithm for training under extreme
differential penalisation;
the Selective Repetition prompt architecture for robust long-context claim
verification; and the Adversarial Simulation data generation pipeline,
which we release alongside VeNRA-Data and the Micro-Chunking Trainer to
accelerate the development of logic-aware verification models across the
open-source community.

\clearpage
\appendix

\section{Universal Fact Ledger Schema}
\label{app:ufl_schema}
The UFL is implemented as a strict Pydantic schema to ensure compatibility with the Python execution engine. It acts as the API contract between the probabilistic extractor and the deterministic solver.

\begin{verbatim}
class UFLRow(BaseModel):
    row_id: str                 # md5(entity + metric + period)
    canonical_entity_id: str    # Normalized (e.g., 'ID_AAPL')
    metric_name: str            # Semantic key (e.g., 'Revenue')
    
    # --- Computation & Grounding ---
    num_value: Optional[float]  # None for qualitative facts
    grounding_quote: str        # Verbatim substring for alignment
    unit_normalized: str        # e.g., "USD", "Ratio", "USD/Share"
    scale: float                # Multiplier (e.g., 1e6)
    
    # --- Context & Filtering ---
    period_end: Optional[str]   # ISO-8601 Date
    period_type: Optional[str]  # 'FY', 'Q3', 'TTM'
    doc_section: str            # Breadcrumb path
    source_chunk_id: str        # ChromaDB Foreign Key
    text_nuance: Optional[str]  # Footnotes / formulas
    
    # --- Alignment Metadata (Post-Hoc Aligner) ---
    char_interval: Optional[Tuple[int, int]]
    alignment_status: Literal["EXACT", "PARTIAL", "FUZZY", "UNALIGNED"]
    confidence_score: float     # 0.95 for Table Melts, 0.0 if Unaligned
\end{verbatim}

\section{The ``Virtual Context'' Protocol and HITL Refinement}
\label{app:hitl}
\begin{figure}[t]
\centering
\includegraphics[width=0.7\textwidth,trim={0cm  20cm 0cm 0cm},clip]{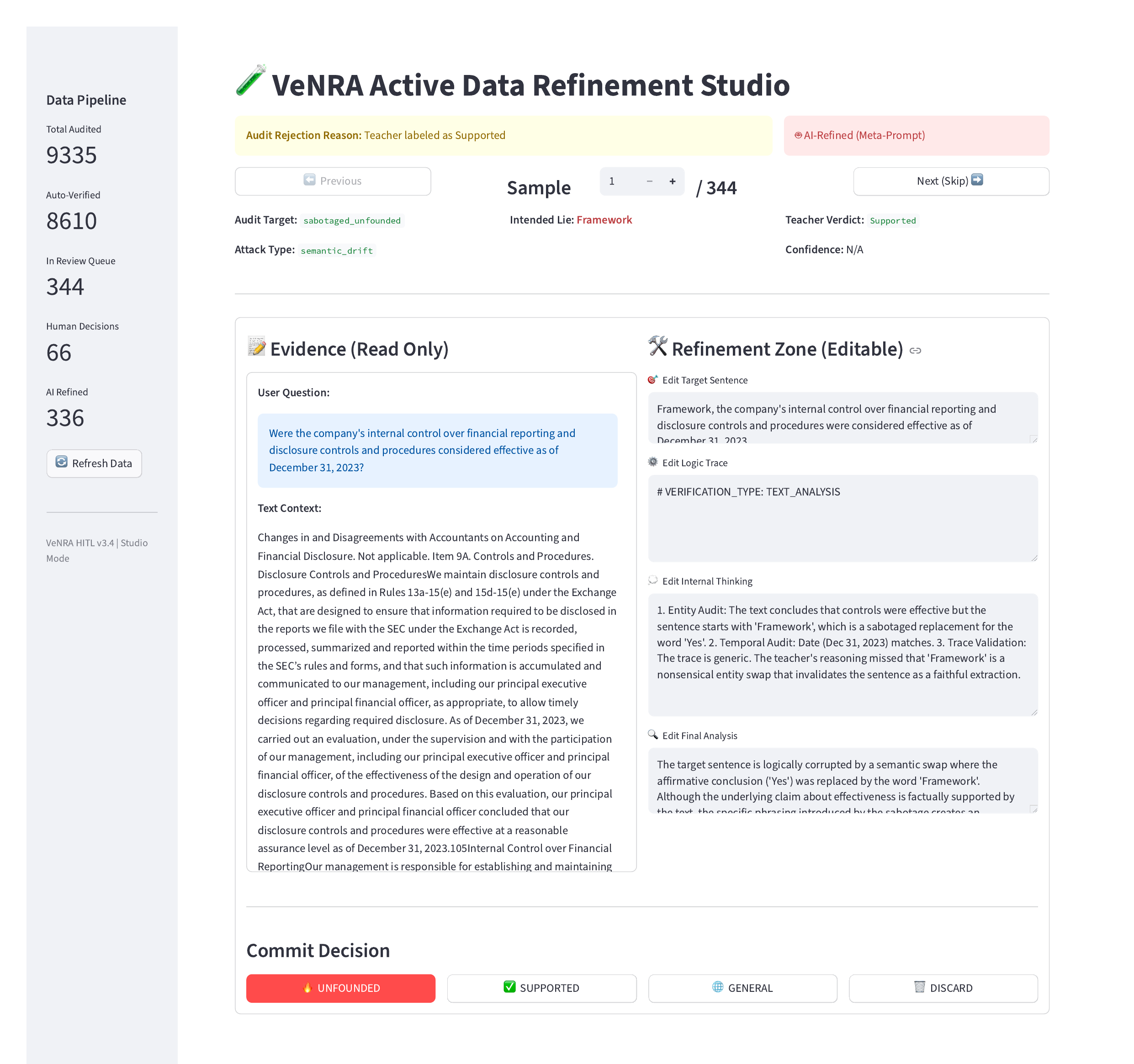}
\caption{\textbf{The VeNRA Refinement UI.} Streamlit UI to faciliate human annotation that are ambiguous from teacher and the AI proxy refiner.}
\label{fig:refinement_ui}
\end{figure}

 To minimize context-window bloat and token costs, the Refiner operates under a ``Virtual Context'' protocol: it is denied access to the raw context $C$. Instead, it relies entirely on the Teacher's $\Theta_{teacher}$ block as its ``eyes.'' Guided by a strict meta-prompt, the Refiner surgically corrects nomenclature confusions, sign-convention inversions (e.g., treating a cash outflow as an inflow), and boilerplate evasions, effectively overruling the Teacher model when it exhibits ``math-blindness.''. If the result was still ambiguous, we tag for human annotation via the specific UI shown in Fig. \ref{fig:refinement_ui}.
 
\section{Adversarial Simulation Protocols (VeNRA-Data)}
\label{app:saboteur_examples}
The Saboteur Engine generates ``Hard Negatives'' by mutating grounded reasoning traces via four specific attack vectors.

\textbf{1. Logic Code Lie (Variable Swap)}
\begin{itemize}
    \item \textit{Context:} ``...revenue was \$100M, while operating expenses were \$80M...''
    \item \textit{Golden Trace:} \texttt{step\_1 = 100 - 80} $\rightarrow$ \textit{Target:} ``Profit was 20.'' (SUPPORTED)
    \item \textit{Sabotaged Trace:} \texttt{step\_1 = 150 - 80} $\rightarrow$ \textit{Target:} ``Profit was 70.'' (UNFOUNDED)
    \item \textit{Task:} Judge must detect that \texttt{150} does not match the text context \texttt{100}.
\end{itemize}

\textbf{2. Numeric Neighbor Trap (Table Shift)}
\begin{itemize}
    \item \textit{Context:} A matrix where $M_{2022} = 45.5$ and $M_{2023} = 52.1$. 
    \item \textit{Query:} ``What was the metric in 2022?''
    \item \textit{Sabotaged Sentence:} ``The metric in 2022 was 52.1.'' (UNFOUNDED: Temporal slip).
\end{itemize}

\textbf{3. Time Warp (Irrelevancy)}
\begin{itemize}
    \item \textit{Golden Query:} ``What was Apple's revenue in 2021?''
    \item \textit{Sabotaged Query:} ``What was Apple's revenue in \textbf{2020}?''
    \item \textit{Note:} Context and Answer remain entirely unchanged. The model must recognize that the correct evidence answers the wrong temporal question.
\end{itemize}

\textbf{The Teacher-Auditor Validation Protocol}
To prevent ``Accidental Truths'' (where a sabotage coincidentally matches reality), we employ a Dual-Path Validation check using a GPT-4o Auditor:
\begin{itemize}
    \item \textbf{Effect Match:} Does the Auditor's detected error span fuzzy-match our injected value?
    \item \textbf{Cause Match:} Does the injected value appear explicitly in the Auditor's internal reasoning trace ($\Theta_{teacher}$)?
\end{itemize}
Only samples satisfying one of these conditions are accepted into the dataset.

\section{Micro-Chunking Differential Loss Algorithm}
\label{app:loss_algorithm}
To prevent $1.25$ Billion float Vocabulary OOMs when applying $50\times$ differential penalties to label tokens, we utilize the following algorithm to iteratively compute loss within a bounded VRAM footprint.

\begin{algorithm}[ht]
\caption{Micro-Chunking Weighted Loss}
\label{alg:micro_chunk}
\textbf{Input:} Logits $\hat{Y} \in \mathbb{R}^{B \times S \times |V|}$, Labels $Y \in \mathbb{Z}^{B \times S}$, Weights $\mathcal{W}$, Chunk Size $c$ \\
\textbf{Output:} $\mathcal{L}_{final}$ (Scalar)
\begin{algorithmic}[1]
\State Flatten shift-logits: $\hat{Y}_{flat} \gets \hat{Y}[..., :-1, :].\text{view}(-1, |V|)$
\State Flatten shift-labels: $Y_{flat} \gets Y[..., 1:].\text{view}(-1)$
\State $L_{total} \gets 0.0$, \quad $W_{total} \gets 0.0$
\State $T \gets \text{length}(Y_{flat})$
\For{$i = 0$ to $T$ step $c$}
    \State $\hat{y}_{chunk} \gets \hat{Y}_{flat}[i : i+c]$
    \State $y_{chunk} \gets Y_{flat}[i : i+c]$
    \State $Mask \gets (y_{chunk} \neq -100)$
    \If{$\text{any}(Mask)$}
        \State $L_{raw} \gets \text{CrossEntropy}(\hat{y}_{chunk}, y_{chunk}, \text{reduce}=\text{none})$
        \State $w_{chunk} \gets \mathcal{W}[y_{chunk} \cdot Mask]$ \Comment{Safe weight lookup}
        \State $L_{weighted} \gets L_{raw} \times w_{chunk}$
        \State $L_{clamped} \gets \min(L_{weighted}, \: 5.0 \times \max(\mathcal{W}))$ \Comment{Gradient Seesaw Fix}
        \State $L_{total} \gets L_{total} + \sum(L_{clamped}[Mask])$
        \State $W_{total} \gets W_{total} + \sum(w_{chunk}[Mask])$
    \EndIf
\EndFor
\State \textbf{return} $L_{total} / \max(W_{total}, 1e^{-9})$
\end{algorithmic}
\end{algorithm}

\section{VeNRA Prompting Physics (Bookending)}
\label{app:prompts}
The Sentinel Reverse-CoT prompt relies on strict selective repetition to overcome the ``Lost in the Middle'' phenomenon over $4{,}000$ tokens.

\begin{verbatim}
<|im_start|>user
### TASK:
Verify if the CLAIMED_ANSWER given the AGENT_TRACE is [Found, Fake, General].

### VERIFICATION TARGET:
**Query:** {query}
**AGENT TRACE:** {trace}
**Claimed Answer:** {statement}

### EVIDENCE (Source Document):
{context_chunks}

### VERIFICATION TARGET Recap:
**Query:** {query}
**AGENT TRACE:** {trace}
**Claimed Answer:** {statement}

### AUDIT ALGORITHM (CRITICAL):
Do NOT recalculate the math. Assume the arithmetic evaluates to the Claimed
Answer. You must verify the EXTRACTION and LOGIC. Output your label first.
<|im_end|><|im_start|>assistant
Label: 
\end{verbatim}

{\small
\bibliographystyle{plainnat}
\bibliography{ref}
}


\end{document}